\documentclass{article}

\usepackage{PRIMEarxiv}

\usepackage[utf8]{inputenc} 
\usepackage[T1]{fontenc}    
\usepackage{hyperref}       
\usepackage{url}            
\usepackage{booktabs}       
\usepackage{amsfonts}       
\usepackage{nicefrac}       
\usepackage{microtype}      
\usepackage{lipsum}
\usepackage{fancyhdr}       
\usepackage{graphicx}       
\graphicspath{{media/}}     

\usepackage{mathtools}
\usepackage{amsmath}
\usepackage[T1]{fontenc}
\usepackage{makecell}
\usepackage[table]{xcolor}
\usepackage{multirow}

\title{Parallel Vision Token Scheduling for Fast and Accurate Multimodal LMMs Inference
\thanks{\textit{\underline{Citation}}: 
\textbf{Authors. Title. Pages.... DOI:000000/11111.}} 
}

\author{
  Wengyi Zhan \\
  Xiamen University \\
  Xiamen, China\\
  \texttt{zhanwy@stu.xmu.edu.cn} \\
   \And
  Mingbao Lin \\
  Rakuten Asia Pte. Ltd. \\
  Singapore \\
  \texttt{linmb001@outlook.com} \\
  \And
Zhihang Lin \\
  Xiamen University, China \\
  Shanghai Innovation Institute, China\\
  \texttt{ lzhedu@foxmail.com} \\
  \And
  Rongrong Ji \\
  Xiamen University \\
  Xiamen, China \\
  \texttt{rrji@xmu.edu.cn} \\
}

\begin{document}
\maketitle

\begin{abstract}
    Multimodal large language models (MLLMs) deliver impressive vision‑language reasoning, but suffer steep inference latency because self‑attention scales quadratically with sequence length and thousands of visual tokens contributed by high‑resolution images. Naively pruning less-informative visual tokens reduces this burden, yet indiscriminate removal can strip away contextual cues essential for background or fine‑grained questions, undermining accuracy.
In this paper, we present ParVTS (Parallel Vision Token Scheduling), a training-free scheduling framework that partitions visual tokens into subject and non-subject groups, processes them in parallel to transfer their semantics into question tokens, and discards the non-subject path mid-inference to reduce computation. This scheduling reduces computational complexity, requires no heuristics or additional modules, and is compatible with diverse existing MLLM architectures.
Experiments across multiple MLLM backbones show that ParVTS prunes up to 88.9\% of visual tokens with minimal performance drop, achieving 1.77× speedup and 70\% FLOPs reduction.
\end{abstract}


\section{Introduction}
\label{sec:intro}

Multimodal large language models (MLLMs)~\cite{liu2024improvedllava,li2023blip2,wu2024towards} with large language models (LLMs)~\cite{radford2021clip,dosovitskiy2020vit}, which are fine-tuned for instruction following, have significantly enhanced capabilities in vision-language tasks, including complex reasoning and visual understanding. However, these benefits come at a steep computational cost.

A primary source of this cost is the quadratic complexity of the transformer's self-attention mechanism~\cite{vaswani2017attention}, which becomes prohibitive as input sequence length increases. In MLLMs, visual tokens from high-resolution images often dominate the sequence---sometimes numbering in the thousands---far exceeding textual tokens. This imbalance greatly prolongs inference latency, posing serious challenges in latency-sensitive applications such as visual question answering (VQA) and mobile augmented reality, where real-time response is required~\cite{card2018psychology}.

To address this issue, prior works~\cite{fastv,bolya2022tome,vtw,prumerge,alvar2025divprune} have proposed reducing redundant visual tokens. A key observation is that only a small subset---often referred to as \textit{subject tokens}---is necessary to answer most visual questions. Retaining these critical tokens allows inference with reduced complexity $\mathcal{O}(L_s^2)$ compared to $\mathcal{O}(L^2)$ for the full sequence, where $L_s \ll L$.

\begin{figure*}[!t]
    \centering
    \includegraphics[width=\textwidth]{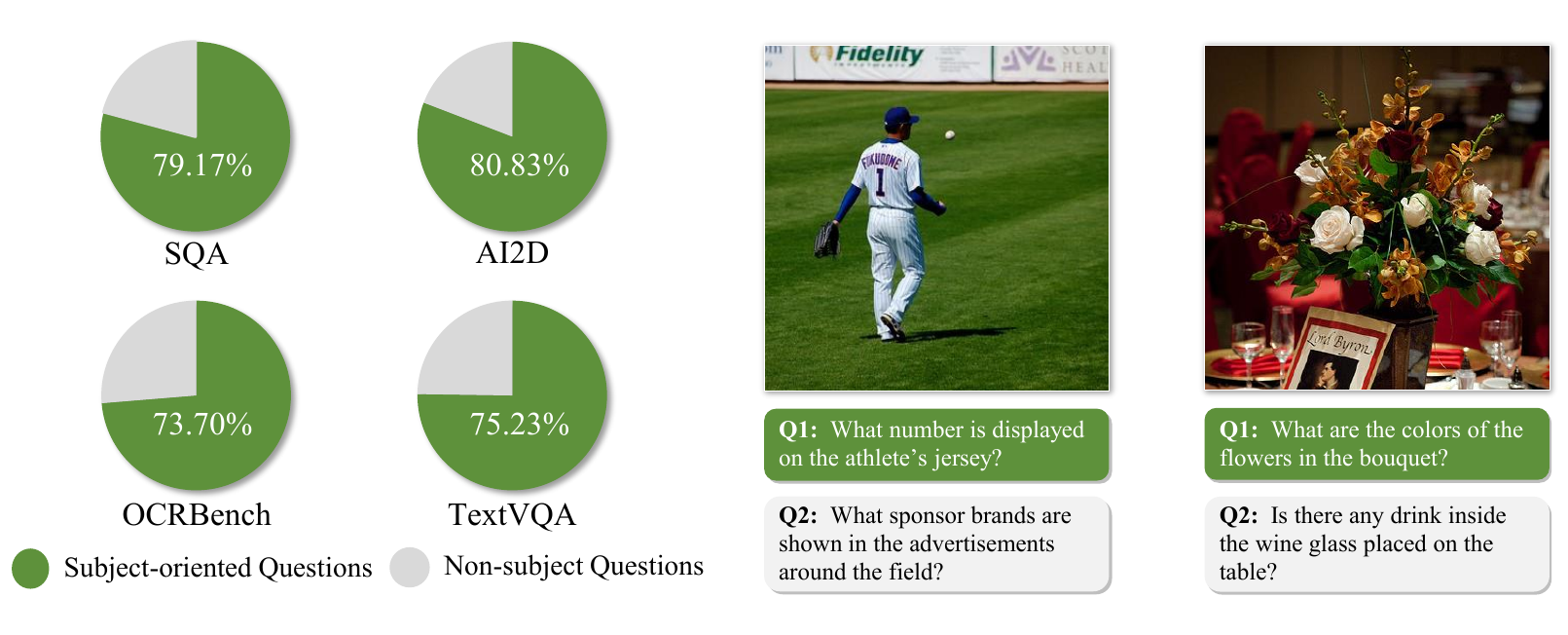}
    \vspace{-1.0em}
    \caption{
    Subject-oriented question distribution and examples across VQA Datasets~\cite{lu2022sqa,liu2024ocrbench,Singh_2019_CVPR_textvqa,ai2d}.
    Left: Percentage of questions focused on subject content, with annotation details in Appendix\,\ref{sec:subject_annotation}.
    Right: Visual examples contrasting subject-relevant (Q1) and non-subject (Q2) questions.
    }
    \label{fig:percentage}
    \vspace{-1.0em}
\end{figure*}

Nonetheless, while a majority of visual questions center around subject entities, a non-negligible portion of queries target background context, fine-grained details, or peripheral objects---elements often represented by \textit{non-subject tokens}. As illustrated in Figure\,\ref{fig:percentage}, subject-related questions account for approximately 73\% to 80\% of all queries across four representative VQA datasets (SQA~\cite{lu2022sqa}, AI2D~\cite{ai2d}, OCRBench~\cite{liu2024ocrbench}, and TextVQA~\cite{Singh_2019_CVPR_textvqa}), leaving 19\% to 27\% as non-subject-oriented. These non-subject questions, though fewer, often require reasoning about subtle attributes (\emph{e.g.}, brand names on signage or the presence of transparent objects) that lie outside the core subject region.

The visual examples on the right of Figure\,\ref{fig:percentage} highlight this distinction. Identifying the jersey number (Q1) requires only subject tokens, whereas recognizing the sponsor logos around the field (Q2) depends on peripheral visual information. While flower color (Q1) is localized to the subject, checking whether a glass contains liquid (Q2) requires attending to less salient yet relevant regions.
These findings underscore that pruning non-subject tokens risks omitting critical visual cues. Thus, a mechanism that reuses or preserves information from these tokens is essential for maintaining both computational efficiency and robust task performance across diverse multimodal scenarios.

Existing methods fall into two main categories. \textbf{(1) Training-free approaches} such as PruMerge~\cite{prumerge} and SparseVLM~\cite{zhang2024sparsevlm}, merge or prune tokens based on similarity heuristics, but lose access to original token representations and may struggle with task generalization. \textbf{(2) Training-based methods} like LLaVA-Mini~\cite{zhang2025llavamini} and VoCo-LLaMA~\cite{ye2025vocollama} introduce extra modules to compress visual information before reduction, which adds training and inference overhead and risks losing fine details.

We argue that an ideal solution should satisfy three criteria: (i) \textbf{computational efficiency} below $\mathcal{O}(L^2)$; (ii) \textbf{heuristic-free reuse} of discarded token information; and (iii) \textbf{structural compatibility} with current MLLMs without adding extra modules.

Recent studies~\cite{vtw, yin2024himap} have highlighted a phenomenon we term \textit{visual information migration}: in early LLM layers, visual token information is implicitly transferred to question tokens via self-attention. This observation motivates a new paradigm: instead of compressing or recovering dropped tokens explicitly, can we leverage this migration to distill essential information early in the network?

In this paper, we propose \textit{ParVTS} (Parallel Vision Token Scheduling), a novel token scheduling framework that enables fast and accurate MLLM inference by deliberately decoupling the processing of different visual token types. Rather than treating all visual tokens uniformly, ParVTS partitions them into \textit{subject} and \textit{non-subject} groups based on their attention weights to the [CLS] token in the vision encoder---where higher attention indicates greater semantic relevance to the main visual focus. This soft saliency-based separation reflects each token’s potential contribution to downstream reasoning, and is computed efficiently without additional supervision or model components.

Once grouped, these tokens are routed through parallel LLM pathways in a single forward pass, implemented via batch-wise token scheduling. Each pathway carries its own copy of the question tokens and attends to a distinct subset of visual inputs. During the early transformer layers, the inherent attention dynamics of the model facilitate visual information migration~\cite{vtw,yin2024himap}---where visual tokens, regardless of type, progressively pass their embedded content into the question tokens. This allows each branch to distill the relevant visual context into its question representation over time.

After a fixed number of layers, the two sets of question tokens---now enriched with either subject- or non-subject-related information---are merged. Since each branch has already conveyed the essential portion of its visual semantics through attention, the fused question tokens possess a sufficient understanding of the image to guide subsequent reasoning.
We then discard the non-subject visual branch and continue inference solely with the subject tokens and merged question tokens, thus achieving significant computational savings while preserving task-relevant information.

This design enables early-stage information transfer from all visual tokens while eliminating redundant computation in later layers. Notably, ParVTS requires no auxiliary modules, heuristics, or fine-tuning, and integrates seamlessly into existing MLLM architectures.

We summarize our main contributions as follows:
(1) We introduce a lightweight, inference-time token scheduling framework that reuses non-subject token information without incurring $\mathcal{O}(L^2)$ complexity.
(2) We demonstrate how visual information migration in early transformer layers enables implicit knowledge transfer, allowing us to discard non-subject paths mid-inference with minimal loss.
(3) Experiments on multiple MLLM backbones show that ParVTS prunes up to 88.9\% of visual tokens while maintaining performance, achieving up to 1.77$\times$ speedup and reducing FLOPs by 70\%.

\section{Related Work}
\label{sec:related_work}

\subsection{Multimodal Large Language Models}
MLLMs extend traditional language models by incorporating additional modalities such as vision and audio, excelling at VQA and multimodal reasoning~\cite{liu2024improvedllava,li2023blip2,zhu2023minigpt4,liu2023llava}. A typical MLLM architecture consists of a vision encoder and a language model, using lightweight modules for alignment, such as MLPs, Q-Formers, or Resamplers~\cite{liu2024improvedllava,zhu2023minigpt4,alayrac2022flamingo}. Representative models include LLaVA~\cite{liu2024improvedllava}, the BLIP family~\cite{li2023blip2, blip3xue2024xgen}, and mini-Gemini-HD~\cite{li2024minigemini}, which integrate CLIP~\cite{radford2021clip} or ViT~\cite{dosovitskiy2020vit} with language models like LLaMA~\cite{touvron2023llama}, GPT~\cite{ouyang2022training,achiam2023gpt} or Gemma-3~\cite{team2025gemma3}. These models employ fine-tuning or frozen strategies to enable image-to-text generation and cross-modal alignment. Furthermore, recent advancements have expanded MLLMs to video and audio understanding, as exemplified by Video-LLaVA~\cite{lin2023videollava} and VideoPoet~\cite{kondratyuk2023videopoet}.

A key challenge in MLLMs is their reliance on encoding images or videos into hundreds or even thousands of visual tokens, which are then concatenated with text tokens and jointly processed by the language model. This approach incurs high computational cost due to the quadratic complexity of self-attention~\cite{vaswani2017attention}. Moreover, the redundancy and low information density of these visual tokens---particularly in high-resolution or multi-frame inputs, as seen in LLaVA~\cite{liu2024improvedllava} and mini-Gemini-HD~\cite{li2024minigemini}---have emerged as significant bottlenecks, significantly affecting inference efficiency.

\subsection{Visual Token Compression}
The problem of visual token redundancy has been studied in the context of vision transformers (ViTs)~\cite{dosovitskiy2020vit}. For example, CF-ViT~\cite{chen2023cfvit} adopts a coarse-to-fine processing strategy, while Evo-ViT~\cite{xu2022evovit} introduces an adaptive slow-fast token evolution mechanism to reduce redundant computation and improve inference efficiency.
In MLLMs, the computational burden caused by excessive visual tokens is even more pronounced, leading to the development of various visual token compression techniques specifically designed to address this issue. FastV~\cite{fastv} selects the most important tokens based on attention scores, retaining only critical information to reduce processing overhead. PruMerge~\cite{prumerge} adaptively prunes and merges tokens by measuring their similarity with class tokens, effectively balancing accuracy and efficiency. SparseVLM~\cite{zhang2024sparsevlm} utilizes cross-modal attention to identify and retain the most relevant visual tokens based on text input, thereby improving token selection and overall model efficiency. These methods leverage distinct strategies to identify and preserve key visual tokens, significantly enhancing the efficiency of MLLMs while maintaining strong performance.

\subsection{Visual Information Migration in MLLMs}
As research on the internal mechanisms of MLLMs gains traction, recent studies have explored how visual information propagates through transformer layers within LLMs.
VTW~\cite{vtw} demonstrates that visual information rapidly migrates to question tokens via causal self-attention in early layers, after which the visual tokens become largely redundant, allowing for their removal in later layers to enable more efficient inference.
HiMAP~\cite{yin2024himap} HiMAP~\cite{yin2024himap} proposes a staged migration process: in shallow layers, visual tokens inject information into question tokens, while in middle layers, they mainly engage in intra-visual aggregation, suggesting a transition from cross-modal fusion to intra-modal consolidation.
Cross-modal Information Flow~\cite{zhang2024crossinformationflow} refines this understanding by identifying two distinct phases of visual-to-text transfer: an initial injection of global visual semantics into question tokens, followed by a more focused transfer of task-relevant regional features. Ultimately, the final predictions rely on the transformed textual representations.

\section{Methodology}
\label{sec:method}

\subsection{Preliminary Observation and Motivation}\label{sec:preliminary}
Modern MLLMs, such as LLaVA~\cite{liu2023llava}, typically consist of three core components: a vision encoder, a cross-modal projector, and a pre-trained LLM. The vision encoder (\emph{e.g.}, CLIP ViT-L~\cite{radford2021clip}) extracts visual patch features and maps them into the language embedding space via a projector, yielding visual tokens aligned with text representations.
Given a multimodal input, the system encodes the task instruction (\emph{i.e.}, the system prompt), the user query, and visual tokens. These are tokenized into system tokens, textual tokens, and visual tokens, respectively. During autoregressive decoding, previously generated outputs are appended to the input sequence. At the first transformer layer ($i=1$), the full input is formulated as:
\begin{equation}\label{eq:concat}
X^t_1 = \mathrm{concat}(S_1, V_1, T_1, O^t_1),
\end{equation}
where $S_1$, $V_1$, and $T_1$ represent the embeddings of the system prompt, visual input, and user query, respectively; $O^t_1$ denotes the generated output tokens at time step $t$, with $O^0_1 = \emptyset$ when $t = 0$.
For subsequent layers ($i > 1$), the hidden representations are updated through a stack of $N$ transformer layers as:
\begin{equation}
X^t_i = \mathrm{TransformerLayer}_{i}(X^t_{i-1}), \quad \text{for } i = 2, \dots, N.
\end{equation}
At each decoding step $t$, the model predicts the next token based on the final hidden representation.

\noindent \textbf{Observation.}
In real-world MLLM applications, user queries typically require grounding in the image content represented by $V_1$. Notably, the number of visual tokens---especially for high-resolution images---often dominates the input sequence length, introducing substantial computational overhead. Empirical observations (see Figure\,\ref{fig:percentage}) reveal that most queries focus on salient foreground entities or regions. This suggests that a subset of visual tokens, denoted $V_1^{\text{sub}} \subseteq V_1$, is often sufficient for correctly answering the majority of vision-grounded questions. 
This motivates an efficiency-oriented reformulation of the input as:
\begin{equation}
X^t_1 = \mathrm{concat}(S_1, V_1^{\text{sub}}, T_1, O^t_1), \quad \text{where } V_1^{\text{sub}} \subseteq V_1.
\end{equation}
To construct $V_1^{\text{sub}}$, we select the top-$k$ visual tokens with the highest attention weights from the [CLS] token in the vision encoder. This strategy is inspired by prior findings~\cite{xu2022evovit,chen2023cfvit}, showing that attention weights to the [CLS] token effectively capture foreground or salient regions in the image, providing a lightweight and supervision-free proxy for visual importance.

Unlike previous works that leverage [CLS]-guided saliency for unimodal vision tasks~\cite{wang2024cls,yu2025sparsity}, we integrate this principle into a saliency-adaptive token scheduling framework for efficient multimodal inference. This enables us to identify subject-centric tokens without additional training, while maintaining compatibility with off-the-shelf MLLMs.

\noindent \textbf{Motivation.}
While this token selection strategy can greatly reduce inference cost, it risks excluding non-subject tokens $V_1^{\text{non}} = V_1 \setminus V_1^{\text{sub}}$, which may carry crucial contextual cues---such as text in the background, transparent objects, or spatial configurations. Figure\,\ref{fig:percentage} shows that approximately 19\%\textasciitilde27\% of real-world visual queries rely on such peripheral information.

This raises a central question: \emph{Can we achieve the efficiency of pruning $V_1^{\text{non}}$, while still benefiting from its semantic contributions?}

To address this, we draw inspiration from the phenomenon of \textit{visual information migration}~\cite{vtw,yin2024himap}, wherein early transformer layers enable semantic transfer from visual tokens to the question tokens via self-attention. This mechanism allows the model to briefly access all visual inputs---both subject and non-subject---and progressively distill their information into a compact question representation. We build on this insight to design a strategy that jointly optimizes computational efficiency and information retention, as detailed below.

\begin{figure*}[t]
    \centering
    \includegraphics[width=\textwidth]{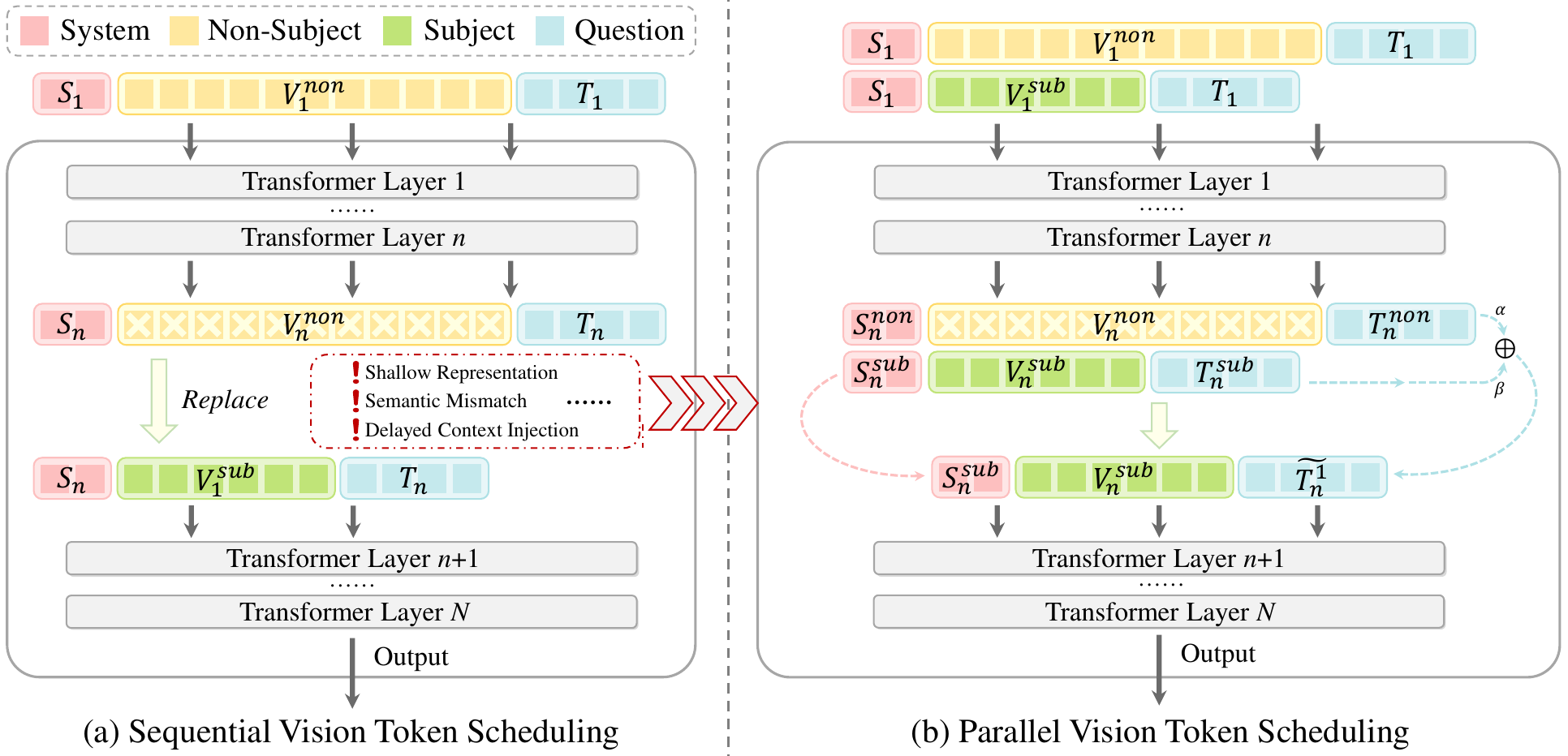}
    \caption{
    The framework of ParVTS. 
    (a) Sequential vision token scheduling (taking non-subject-first as an example) injects token groups at different transformer layers, leading to a series of issues.
    (b) Our parallel scheduling enables both token groups to participate in early layers simultaneously, ensuring sufficient information migration and consistent representation with low inference cost.
    }
    \label{fig:framework}
\end{figure*}

\subsection{Vision Token Scheduling: When and How Visual Tokens are Used}
\label{sec:when_and_how}

Given the partitioned visual token sets $V^{\text{sub}}_1$ and $V^{\text{non}}_1$ described in Sec.\,\ref{sec:preliminary}, we now consider a central question: \emph{When and how should each group of tokens be involved in inference?} Intuitively, both token types carry complementary visual information---subject tokens reflect salient entities, while non-subject tokens encode contextual or background cues. Efficient utilization requires a scheduling strategy that enables both groups to contribute meaningfully without incurring full attention overhead.

To this end, our vision token scheduling temporally separates subject and non-subject tokens across transformer layers. Specifically, we leverage the phenomenon of \textit{visual information migration}, where visual semantics are transferred into question tokens via self-attention in early layers. 
We explore two sequential scheduling strategies: (1) Subject-First Scheduling; and (2) Non-Subject-First Scheduling. 

\noindent \textbf{Strategy A: Subject-First Scheduling.}
A straightforward design is to first use subject tokens $V^{\text{sub}}_1$ to inject object-centric semantics into question stream, and later introduce non-subject tokens $V^{\text{non}}_1$ to provide complementary context.

\noindent \textbf{\textit{Stage 1: Early Subject Integration.}} During the initial $n$ transformer layers, only the subject tokens are included. Through causal attention, question tokens attend to these subject tokens, progressively absorbing salient visual information into their representations:
\begin{equation}
    X^1_{n} = \mathrm{Transformer}_{1:n}(\mathrm{concat}(S_1, V^{\text{sub}}_1, T_1)),
\end{equation}
where $S_1$ and $T_1$ represent the system and question tokens, respectively.

\noindent \textbf{\textit{Stage 2: Late Context Injection}.} At layer $n$, we remove the subject tokens and insert the original embeddings of the non-subject tokens---those not previously involved---to continue inference:
\begin{equation}
\tilde{X}^1_n = \mathrm{ReplaceVision}(X^1_n, V^{\text{sub}}_n \rightarrow V^{\text{non}}_1),
\end{equation}
where $\mathrm{ReplaceVision}(\cdot)$ replaces only the visual token slots while preserving ordering and retaining the hidden states of the non-visual tokens.

This allows non-subject tokens to interact with already-refined question tokens: 
\begin{equation}
\label{eqa:main_branch}
X^1_{n+1} = \mathrm{Transformer}_{n+1}(\tilde{X}^1_n).
\end{equation}
In this stage, attention enables the injection of contextual or peripheral information from $V^{\text{non}}_1$ into the question representation.

\noindent \textbf{\textit{Efficiency and Limitations}.}
This temporal scheduling reduces overall complexity: early layers operate on $|V^{\text{sub}}_1|$ visual tokens, and later layers on $|V^{\text{non}}_1|$, both subsets being much smaller than the full token length $L$. The attention cost drops from $\mathcal{O}(L^2)$ to $\mathcal{O}(L_s^2)$ and $\mathcal{O}((L - L_s)^2)$ in the two respective stages.

While intuitive, relying on a single merge depth $n$ to transition from subject to non-subject tokens inherently induces two trade-offs:
\begin{itemize}
\item \textbf{Insufficient Subject Transfer at Shallow Merge Depths.} 
A smaller merge depth $n$ restricts the semantic migration capacity of subject tokens. Since over 70\% of questions are subject-oriented (Sec.\,\ref{sec:preliminary}), limiting their exposure to only the earliest layers impairs the model’s ability to integrate critical object-level information into the question token representations.
\item \textbf{Delayed Context Injection and Limited Efficiency Gains.} 
Conversely, a larger $n$ delays the introduction of non-subject tokens, which carry crucial context, thereby diminishing their potential contribution.
In addition, under high pruning rates, retaining numerous active non-subject tokens becomes inefficient, as little information is transferred in the later layers.
\end{itemize}

\noindent \textbf{Strategy B: Non-Subject-First Scheduling.}
Alternatively, the order can be reversed by injecting $V^{\text{non}}_1$ in early layers and switching to $V^{\text{sub}}_1$ later, as shown in Figure\,\ref{fig:framework} (a). This variant eases tension in $n$ selection: since less than ~30\% of questions rely on non-subject information, even a shallow $n$ allows sufficient migration.
Moreover, an earlier switch enables more aggressive pruning of visual tokens in later layers, leading to improved efficiency.

Yet, this strategy inherits two fundamental problems:
\begin{itemize}
\item \textbf{Shallow representation.} Subject tokens bypass early transformer layers and thus miss hierarchical refinement, resulting in weakly encoded representations.
\item \textbf{Semantic mismatch.} When injected at layer $n$, subject tokens are processed in isolation from question tokens that already reside in a higher-level latent space, thereby resulting in semantic misalignment and reduced attention effectiveness.
\end{itemize}

To overcome these limitations, we propose a parallel execution scheme in the next section that enables both token groups to participate from the beginning, while still maintaining low inference cost.

\subsection{Parallel Path Execution for Vision Token Groups}
\label{sec:parallel}

To overcome the representational limitations in our vision token scheduling, we propose a \textit{parallel execution} strategy that enables both subject and non-subject tokens to participate in the early transformer layers simultaneously. This ensures comprehensive visual information migration into the question tokens while avoiding the semantic mismatch caused by delayed token injection.

A naive solution would be to process both token groups sequentially, passing each through the same early layers. However, this doubles the compute cost and negates the benefits of scheduling. Instead, we adopt a \textit{batch-parallel execution} design: both visual token groups are processed independently in the same forward pass by concatenating their input sequences along the batch dimension.

As illustrated in Figure\,\ref{fig:framework}(b), we construct two parallel input streams for the first $n$ transformer layers:
\begin{align}
X^{\text{non}}_n \mathrm{Transformer}_{1:n}([S_1, V^{\text{non}}_1, T_1]),\;\; \\
X^{\text{sub}}_n = \mathrm{Transformer}_{1:n}([S_1, V^{\text{sub}}_1, T_1]),
\end{align}
where both sequences share the same system and question tokens but differ in visual content.

Let the outputs be decomposed as:
\begin{equation}
X^{\text{non}}_n = [S^{\text{non}}_n, V_n^{\text{non}}, T^{\text{non}}_n],\;\; 
X^{\text{sub}}_n = [S^{\text{sub}}_n, V_n^{\text{sub}}, T^{\text{sub}}_n].
\end{equation}
We then combine the question token representations using weighted averaging:
\begin{equation}
\label{eq:alpha}
\tilde{T}^1_n = \alpha \cdot T_n^{\text{non}} + \beta \cdot T_n^{\text{sub}},
\end{equation}
yielding the unified state:
\begin{equation}
\tilde{X}^1_n = [S_n^{\text{sub}}, V_n^{\text{sub}}, \tilde{T}^1_n].
\end{equation}
Note that $S_n^{\text{non}} = S_n^{\text{sub}}$ as the system tokens are separately processed in both paths but yield identical representations due to the same inputs and model weights. 
Crucially, to achieve an efficient implementation without duplicating GPU memory for the first $n$ layers, we adopt an optimized engineering approach: we modify the attention mask in the first $n$ layers such that subject and non-subject tokens are mutually invisible, allowing both to flow their information exclusively into the question tokens.
After layer $n$, all non-subject tokens are excluded, and there is no need to store KV caches for them at any layer.
Only the subject visual tokens are retained for subsequent inference, which includes the remainder of the prefilling stage and the entire decoding phase, and they continue from layer $n+1$ following Eq.\,(\ref{eqa:main_branch}).

\noindent  \textbf{Advantages.} This design offers three key benefits:
\begin{itemize}
\item \textbf{Computational efficiency.} Parallel execution avoids duplicated passes while maintaining the $\mathcal{O}(L_s^2)$ complexity in later layers.
\item \textbf{Representation consistency.} Both visual token groups are refined through early transformer layers, avoiding shallow representations and promoting alignment in feature representations.
\item \textbf{Information completeness.} The model benefits from early-stage migration of both subject and non-subject information into the question tokens, preserving essential semantics.
\end{itemize}

\begin{table*}[t]
\renewcommand{\arraystretch}{0.9}
\caption{
Comparative experiments on \textbf{LLaVA-1.5-7B} and \textbf{LLaVA-Next-7B} across multimodal benchmarks. \textbf{Bold} indicates the best result.}
\label{tab:llava1.5-and-llavanext}
\centering
\resizebox{0.8\textwidth}{!}{%
\begin{tabular}{c|cccccccc|c}
\toprule
\multicolumn{1}{c|}{\textbf{Methods}}         & \multicolumn{1}{c}{\textbf{GQA}}           & \multicolumn{1}{c}{\textbf{MMB}}           & \multicolumn{1}{c}{\textbf{MME}}             & \multicolumn{1}{c}{\textbf{POPE}}          & \multicolumn{1}{c}{\textbf{SQA}}           & \multicolumn{1}{c}{\textbf{VQA\textsuperscript{v2}}}                      & \multicolumn{1}{c}{\textbf{VizWiz}}        & \multicolumn{1}{c|}{\textbf{OCRBench}}       & \textbf{Avg.$\uparrow$}                       \\ \hline

\rowcolor[HTML]{EFEFEF} 
\multicolumn{1}{c|}{\cellcolor[HTML]{EFEFEF}LLaVA-1.5-7B~\cite{liu2023llava}} & \multicolumn{8}{c|}{\cellcolor[HTML]{EFEFEF}\textit{All Tokens (576, 100.00\%)}}    & \multicolumn{1}{c}{\cellcolor[HTML]{EFEFEF}}     \\ 

\multicolumn{1}{c|}{Vanilla}    & \multicolumn{1}{c}{61.9}         & \multicolumn{1}{c}{64.1}         & \multicolumn{1}{c}{1866.1}         & \multicolumn{1}{c}{85.9}         & \multicolumn{1}{c}{69.6}         & \multicolumn{1}{c}{78.5}                      & \multicolumn{1}{c}{54.3}         & \multicolumn{1}{c|}{313.0}         & 100.00 \%                       \\ \hline

  \rowcolor[HTML]{EFEFEF} 
\multicolumn{1}{c|}{\cellcolor[HTML]{EFEFEF}} & \multicolumn{8}{c|}{\cellcolor[HTML]{EFEFEF}\textit{Token Reduction  (64, $-$88.89\%)}}  & \multicolumn{1}{c}{\cellcolor[HTML]{EFEFEF}}                                                  \\ 

\multicolumn{1}{c|}{FastV~\cite{fastv}}          & \multicolumn{1}{c}{49.9}          & \multicolumn{1}{c}{54.5}          & \multicolumn{1}{c}{1466.7}          & \multicolumn{1}{c}{70.8}          & \multicolumn{1}{c}{\textbf{69.0}} & \multicolumn{1}{c}{62.0} & \multicolumn{1}{c}{54.2}          & \multicolumn{1}{c|}{172.0}          & 78.64\% 
\\

\multicolumn{1}{c|}{SparseVLM~\cite{zhang2024sparsevlm}}      & \multicolumn{1}{c}{52.7}          & \multicolumn{1}{c}{56.2}          & \multicolumn{1}{c}{1505.0}          & \multicolumn{1}{c}{75.1}          & \multicolumn{1}{c}{62.2}          & \multicolumn{1}{c}{68.2}                       & \multicolumn{1}{c}{50.1}          & \multicolumn{1}{c|}{180.0}          & 78.64\%                    
\\

\multicolumn{1}{c|}{SAINT~\cite{jeddi2025saint}}          & \multicolumn{1}{c}{\textbf{55.5}} & \multicolumn{1}{c}{58.0}          & \multicolumn{1}{c}{1604.2}          & \multicolumn{1}{c}{\textbf{84.5}} & \multicolumn{1}{c}{67.9}          & \multicolumn{1}{c}{67.7}                       & \multicolumn{1}{c}{\textbf{55.4}} & \multicolumn{1}{c|}{240.0}          & 85.69\%                    
\\ 

\multicolumn{1}{c|}{HiRED~\cite{arif2024hired}}          & \multicolumn{1}{c}{54.6}          & \multicolumn{1}{c}{60.2}          & \multicolumn{1}{c}{1599.0}          & \multicolumn{1}{c}{73.6}          & \multicolumn{1}{c}{68.2}          & \multicolumn{1}{c}{69.7}                       & \multicolumn{1}{c}{50.2}          & \multicolumn{1}{c|}{191.0}          & 83.13\%                    
\\

\multicolumn{1}{c|}{ParVTS (Ours)}     & \multicolumn{1}{c}{55.2}          & \multicolumn{1}{c}{\textbf{61.3}} & \multicolumn{1}{c}{\textbf{1739.4}} & \multicolumn{1}{c}{76.9}          & \multicolumn{1}{c}{68.9}          & \multicolumn{1}{c}{\textbf{70.7}}                       & \multicolumn{1}{c}{55.1}          & \multicolumn{1}{c|}{\textbf{282.0}} & \textbf{92.45\%} 
\\
\hline

\multicolumn{1}{c|}{\cellcolor[HTML]{EFEFEF}LLaVA-Next-7B~\cite{liu2024improvedllava}} & \multicolumn{8}{c|}{\cellcolor[HTML]{EFEFEF}\textit{All Tokens (2880, 100.00\%)}}    & \multicolumn{1}{c}{\cellcolor[HTML]{EFEFEF}}     \\ 

\multicolumn{1}{c|}{Vanilla}    & \multicolumn{1}{c}{62.5}         & \multicolumn{1}{c}{66.2}         & \multicolumn{1}{c}{1875.1}         & \multicolumn{1}{c}{87.8}         & \multicolumn{1}{c}{68.0}         & \multicolumn{1}{c}{79.5}                      & \multicolumn{1}{c}{59.7}         & \multicolumn{1}{c|}{510.0}         & 100.00 \%                       \\ \hline

  \rowcolor[HTML]{EFEFEF} 
\multicolumn{1}{c|}{\cellcolor[HTML]{EFEFEF}} & \multicolumn{8}{c|}{\cellcolor[HTML]{EFEFEF}\textit{Token Reduction  (320, $-$88.89\%)}}  & \multicolumn{1}{c}{\cellcolor[HTML]{EFEFEF}}                                                                          \\ 
\multicolumn{1}{c|}{FastV~\cite{fastv}}               & \multicolumn{1}{c}{55.9}          & \multicolumn{1}{c}{61.6}          & \multicolumn{1}{c}{1661.0}          & \multicolumn{1}{c}{71.7}          & \multicolumn{1}{c}{62.8}          & \multicolumn{1}{c}{71.9}                       & \multicolumn{1}{c}{53.1}          & \multicolumn{1}{c|}{374.0}          &   85.87\%                        
\\ 

\multicolumn{1}{c|}{SparseVLM~\cite{zhang2024sparsevlm}}           & \multicolumn{1}{c}{56.1}          & \multicolumn{1}{c}{60.6}          & \multicolumn{1}{c}{1533.0}          & \multicolumn{1}{c}{82.4}          & \multicolumn{1}{c}{66.1}          & \multicolumn{1}{c}{71.5}                       & \multicolumn{1}{c}{52.0}          & \multicolumn{1}{c|}{270.0}          &    78.03\%
   
 \\ 

\multicolumn{1}{c|}{GlobalCom\textsuperscript{2}~\cite{liu2025globalcom}}           & \multicolumn{1}{c}{57.1}          & \multicolumn{1}{c}{61.8}          & \multicolumn{1}{c}{1698.0}          & \multicolumn{1}{c}{83.8} & \multicolumn{1}{c}{\textbf{67.4}}          & \multicolumn{1}{c}{\textbf{76.7}}              & \multicolumn{1}{c}{54.6}          & \multicolumn{1}{c|}{375.0}          &    88.09\%                        
\\ 

\multicolumn{1}{c|}{HiRED~\cite{arif2024hired}}               & \multicolumn{1}{c}{59.3}          & \multicolumn{1}{c}{\textbf{64.2}} & \multicolumn{1}{c}{1690.0} & \multicolumn{1}{c}{83.3}          & \multicolumn{1}{c}{66.7}          & \multicolumn{1}{c}{75.7}                       & \multicolumn{1}{c}{54.2}          & \multicolumn{1}{c|}{\textbf{404.0}} &      88.91\%
                      
\\

\multicolumn{1}{c|}{ParVTS (Ours)}                & \multicolumn{1}{c}{\textbf{61.2 }} & \multicolumn{1}{c}{63.6 }          & \multicolumn{1}{c}{\textbf{1761.6 }}          & \multicolumn{1}{c}{\textbf{84.8} }          & \multicolumn{1}{c}{67.1 }          & \multicolumn{1}{c}{76.0} & \multicolumn{1}{c}{\textbf{58.5 }} & \multicolumn{1}{c|}{329.0}          & \textbf{89.07\%} 

\\ \bottomrule
\end{tabular}
}
\vspace{-6pt}
\end{table*}

\section{Experiments}

\subsection{Experimental Setup}
\label{sec:setup}
We verify our ParVTS using LLaVA-1.5~\cite{liu2023llava}, LLaVA-Next~\cite{liu2024improvedllava}, InternVL2.5~\cite{chen2024internvl25}, Qwen2.5-VL~\cite{bai2025qwen2}, and Video-LLaVA~\cite{lin2023videollava}. Comparisons are conducted across diverse benchmarks, encompassing: visual question answering (GQA~\cite{hudson2019gqa}, VQAv2~\cite{goyal2017vqav2}, ScienceQA~\cite{lu2022sqa}, VizWiz-VQA~\cite{bigham2010vizwiz}, MMB~\cite{liu2024mmbench}, MME~\cite{mme}), hallucination detection (POPE~\cite{Li-hallucination-2023}), video question answering (TGIF-QA~\cite{jang2017tgif}, MSVD-QA~\cite{xu2017video}).
Additional implementation details are provided in {Appendix\,\ref{sec:implementation}.

\begin{table*}[]
\renewcommand{\arraystretch}{0.9}
\caption{
Comparative experiments on \textbf{InternVL2.5-2B}. \textbf{Bold} indicates the best result.
}
\label{tab:internvl2.5-2b}
\scriptsize
\centering
\resizebox{0.8\textwidth}{!}{%
\begin{tabular}{c|cccccc|c}
\toprule
\multicolumn{1}{c|}{\textbf{Method}} & \multicolumn{1}{c}{\textbf{GQA}}  & \multicolumn{1}{c}{\textbf{MMB}} & \multicolumn{1}{c}{\textbf{MME}} & \multicolumn{1}{c}{\textbf{POPE}} & \multicolumn{1}{c}{\textbf{SQA}}  & \textbf{AI2D} & \multicolumn{1}{c}{\textbf{Avg. $\uparrow$}}         \\ \hline

\rowcolor[HTML]{EFEFEF} 
InternVL2.5-2B~\cite{chen2024internvl25}                       & \multicolumn{6}{c|}{\cellcolor[HTML]{EFEFEF}\textit{All Tokens (Dynamic, 100.00\%)}}                                                                                                                   & \multicolumn{1}{l}{\cellcolor[HTML]{EFEFEF}\textit{}} \\

Vanilla     & \multicolumn{1}{c}{59.5}          & 74.2                              & 2097.9                            & 90.7                               & \multicolumn{1}{c}{96.1}          & 73.87         & 100.00\%                                              \\ \hline

\rowcolor[HTML]{EFEFEF} 
& \multicolumn{6}{c|}{\cellcolor[HTML]{EFEFEF}\textit{Dynamic Token Reduction ($-$88.89\%)}}                                     &                                                       \\

FastV~\cite{fastv}                                 & \multicolumn{1}{c}{\textbf{49.9}} & 64.1                              & 1931.3                            & 84.4                               & \multicolumn{1}{c}{85.1}          & 64.0          & 91.43\%                                               \\

ParVTS (Ours)                         & \multicolumn{1}{c}{44.2}          & \textbf{70.7}                     & \textbf{2027.9}                   & \textbf{85.9}                      & \multicolumn{1}{c}{\textbf{93.2}} & \textbf{67.8} & \textbf{95.88\%}      

\\ \bottomrule
\end{tabular}
}
\vspace{-4pt}
\end{table*}

\begin{table*}[!]
\renewcommand{\arraystretch}{0.9}
\centering
\caption{Comparative experiments on \textbf{Qwen2.5-VL-3B}. \textbf{Bold} indicates the best result.
}
\label{tab:qwen25-3b}
\resizebox{0.8\textwidth}{!}{%
\begin{tabular}{c|ccccccc|c}
\toprule
\multicolumn{1}{c|}{\textbf{Method}} & \multicolumn{1}{c}{\textbf{GQA}} & \multicolumn{1}{c}{\textbf{MMB}} & \multicolumn{1}{c}{\textbf{MME}} & \multicolumn{1}{c}{\textbf{POPE}} & \multicolumn{1}{c}{\textbf{SQA}} & \multicolumn{1}{c}{\textbf{OCRBench}} & \multicolumn{1}{c|}{\textbf{VizWiz}} & \multicolumn{1}{c}{\textbf{Avg. $\uparrow$}} \\ \hline
\rowcolor[HTML]{EFEFEF} 
\multicolumn{1}{c|}{Qwen2.5-VL-3B~\cite{bai2025qwen2}}                        & \multicolumn{7}{c|}{\cellcolor[HTML]{EFEFEF}\textit{All Tokens (Dynamic, 100.00\%)}}         &        \\

\multicolumn{1}{c|}{Vanilla }          & 60.0                              & 78.6                              & 2181.7                            & 87.7                               & 50.6                              & 775.0                                  &\multicolumn{1}{c|}{ 68.5}                                 & \multicolumn{1}{c}{100.00\%   }                                   \\ \hline

\rowcolor[HTML]{EFEFEF} \multicolumn{1}{c|}{} & \multicolumn{7}{c|}{\cellcolor[HTML]{EFEFEF}\textit{Dynamic Token Reduction ($-$88.89\%)}}    &                                               \\

\multicolumn{1}{c|}{FastV~\cite{fastv}}                               & 47.7                              & 59.7                              & 1676.4                            & 74.9                               & 42.7                              & 264.0                                  & \multicolumn{1}{c|}{62.2 }                                & 67.46\%                                       \\

\multicolumn{1}{c|}{ParVTS (Ours) }                        & \textbf{53.5}                     & \textbf{71.1}                     & \textbf{1870.4}                   & \textbf{81.8}                      & \textbf{65.2}                     & \textbf{345.0}                         & \multicolumn{1}{c|}{\textbf{62.5}  }                      & \textbf{77.21\%}                \\  \bottomrule
\end{tabular}
}
\vspace{-4pt}
\end{table*}

%
\begin{table}[]
\renewcommand{\arraystretch}{1.0}
\caption{
Comparative experiments on \textbf{Video-LLaVA-7B}. \textbf{Bold} indicates the best result.
}
\label{tab:videollava}
\scriptsize
\centering
\resizebox{0.7\textwidth}{!}{%
\begin{tabular}{ccccccc}
\toprule
\multicolumn{1}{c|}{}                          & \multicolumn{2}{c|}{\textbf{TGIF}}                                  & \multicolumn{2}{c|}{\textbf{MSVD}}                                  & \multicolumn{2}{c}{\textbf{Avg.$\uparrow$}}                                  \\ \cline{2-7} 
\multicolumn{1}{c|}{\multirow{-2}{*}{\textbf{Methods}}} & \multicolumn{1}{c|}{\textbf{Accuracy}} & \multicolumn{1}{c|}{\textbf{Score}} & \multicolumn{1}{c|}{\textbf{Accuracy}} & \multicolumn{1}{c|}{\textbf{Score}} & \multicolumn{1}{c|}{\textbf{Accuracy}} & \multicolumn{1}{c}{\textbf{Score}} \\ \hline
\rowcolor[HTML]{EFEFEF} 
\multicolumn{1}{c|}{Video-LLaVA-7B~\cite{lin2023videollava} }                                  & \multicolumn{4}{c|}{\cellcolor[HTML]{EFEFEF}\textit{All Tokens (2056, 100.00\%)}}                                                & \textit{}                    & \textit{}                                                       \\

\multicolumn{1}{c|}{Vanilla   }                                      & 47.0                          & 3.4                        & 70.2                          & \multicolumn{1}{c|}{3.9 }                       & 58.2                          & 3.6                        \\  \hline

\rowcolor[HTML]{EFEFEF}    \multicolumn{1}{c|}{ }
  & \multicolumn{4}{c|}{\cellcolor[HTML]{EFEFEF}\textit{Token Reduction (1032, $-$49.80\%)}}                                          & \textit{}                     & \textit{}                                                                                                              \\
\multicolumn{1}{c|}{FastV~\cite{fastv} }                                          & \textbf{45.2 }                         & \textbf{3.1}                        & \textbf{71.0 }                         & \multicolumn{1}{c|}{\textbf{3.9} }                       & \textbf{58.1}                       & \textbf{3.5}                      \\

\multicolumn{1}{c|}{ParVTS (Ours) }                                  & 45.0                           & \textbf{3.1}                         & 70.5                          & \multicolumn{1}{c|}{\textbf{3.9}}                       & 57.8                           & \textbf{3.5}              
\\ \bottomrule
\end{tabular}
}
\vspace{-14pt}
\end{table}

\subsection{Main Results}

\noindent \textbf{Quantitative Evaluation.}
Tables\,\ref{tab:llava1.5-and-llavanext} \textasciitilde~\,\ref{tab:videollava} present the quantitative results of our ParVTS under a fixed vision token budget across multimodal understanding tasks. In Table\,\ref{tab:llava1.5-and-llavanext}, on LLaVA-1.5-7B, our ParVTS achieves an average performance of 92.45\%, ranking first and outperforming the second-best (SAINT~\cite{jeddi2025saint}) by 6.76\%.  
This shows that ParVTS well mitigates the performance degradation caused by token reduction, maintaining strong robustness and stability even under aggressive compression.

Also, our ParVTS improves performance across other open-source MLLMs, as shown in Table\,\ref{tab:internvl2.5-2b}$\sim$\,\ref{tab:videollava}.  
This indicates that ParVTS transfers reliably to diverse model architectures and scales, including InternVL2.5 and Qwen2.5-VL.
Further experiments covering additional vision token budgets, more MLLM architectures (InternVL2, Qwen3-VL), and diverse model sizes are provided in {Appendix\,\ref{sec:more}}.

\noindent \textbf{Qualitative Examples.}
We further provide several inference examples in Appendix\,\ref{sec:visual_cls_token}, showing that ParVTS correctly answers queries related to non-subject tokens. By leveraging non-subject tokens instead of discarding them, our method preserves contextual cues and enables accurate reasoning beyond subject-only regions.

\noindent \textbf{Downstream Task.}
We validate the generalization of ParVTS on LISA~\cite{lai2024lisa}, a fine-grained segmentation task. Results (detailed in Appendix\,\ref{sec:lisa}) show that ParVTS successfully preserves the original model's segmentation capability across diverse reasoning scenarios, confirming its effectiveness in downstream applications requiring detailed visual understanding.

\begin{table*}[!t]
\caption{Comparison of token budget, inference speedup, and computational cost under comparable MME scores using LLaVA-1.5-7B. \textbf{Bold} indicates the best result in each group.}
\label{tab:latency}
\renewcommand{\arraystretch}{0.9}
\scriptsize
\centering
\resizebox{0.8\textwidth}{!}{%
\begin{tabular}{c|cccccc}
\toprule
\multicolumn{1}{c|}{\textbf{Methods}} & \multicolumn{1}{c}{\textbf{\#Tokens $\downarrow$}} & \multicolumn{1}{c}{\textbf{\makecell{Time / Sample \\  (ms) $\downarrow$}}} & \multicolumn{1}{c}{\textbf{Speedup   $\uparrow$}} & \multicolumn{1}{c}{\textbf{TFLOPs   $\downarrow$}} & \multicolumn{1}{c}{\textbf{ \makecell{TFLOPs \\ Ratio   $\downarrow$}}} & \textbf{MME Score   $\uparrow$} \\ \hline

\multicolumn{1}{c|}{LLaVA-1.5-7B~\cite{liu2023llava}}    & \multicolumn{1}{c}{576}                                                               & \multicolumn{1}{c}{285.17}                                     & \multicolumn{1}{c}{1.00$\times$}                          & \multicolumn{1}{c}{8.48}                           & \multicolumn{1}{c}{100.00\%}                             & 1866.10                         \\ \hline

\rowcolor[HTML]{EFEFEF}
\multicolumn{1}{c|}{\cellcolor[HTML]{EFEFEF}}  &
\multicolumn{6}{c}{\cellcolor[HTML]{EFEFEF}\textit{Comparison under similar MME scores ($\approx$1810)}}                                                                                                             \\ 
\multicolumn{1}{c|}{FastV~\cite{fastv}}  & \multicolumn{1}{c}{221}                                                         & \multicolumn{1}{c}{224.94}                                     & \multicolumn{1}{c}{1.27$\times$}                          & \multicolumn{1}{c}{4.63}                           & \multicolumn{1}{c}{54.60\%}                              & 1812.67                         \\ 
\multicolumn{1}{c|}{SparseVLM~\cite{zhang2024sparsevlm}}    & \multicolumn{1}{c}{210}                                                           & \multicolumn{1}{c}{210.19}                                     & \multicolumn{1}{c}{1.36$\times$}                          & \multicolumn{1}{c}{5.85}                           & \multicolumn{1}{c}{68.94\%}                              & 1810.34                         \\ 
\multicolumn{1}{c|}{ParVTS (Ours)}            & \multicolumn{1}{c}{\textbf{161}}                                                          & \multicolumn{1}{c}{\textbf{185.34}}                                     & \multicolumn{1}{c}{\textbf{1.54$\times$}}                          & \multicolumn{1}{c}{\textbf{3.76}}                           & \multicolumn{1}{c}{\textbf{44.34\%}}                              & \textbf{1814.03  }                       \\ \hline

\multicolumn{1}{c|}{\cellcolor[HTML]{EFEFEF}}  &
\multicolumn{6}{c}{\cellcolor[HTML]{EFEFEF}\textit{Comparison under similar MME scores ($\approx$1750)}}   \\
\multicolumn{1}{c|}{FastV~\cite{fastv}}  & \multicolumn{1}{c}{175}                                                          & \multicolumn{1}{c}{188.71}                                     & \multicolumn{1}{c}{1.51$\times$}                          & \multicolumn{1}{c}{4.10}                           & \multicolumn{1}{c}{48.35\%}                              & 1752.60                         \\ 
\multicolumn{1}{c|}{SparseVLM~\cite{zhang2024sparsevlm}}    & \multicolumn{1}{c}{140}                                                         & \multicolumn{1}{c}{193.77}                                     & \multicolumn{1}{c}{1.47$\times$}                          & \multicolumn{1}{c}{5.45}                           & \multicolumn{1}{c}{64.23\%}                              & 1749.23                         \\ 

\multicolumn{1}{c|}{ParVTS (Ours)}            & \multicolumn{1}{c}{\textbf{103}}                                                         & \multicolumn{1}{c}{\textbf{173.55}}                                     & \multicolumn{1}{c}{\textbf{1.64$\times$}}                          & \multicolumn{1}{c}{\textbf{3.16}}                           & \multicolumn{1}{c}{\textbf{37.26\%}}                              & \textbf{1774.10   }                      \\ \hline

 \multicolumn{1}{c|}{\cellcolor[HTML]{EFEFEF}}  &
\multicolumn{6}{c}{\cellcolor[HTML]{EFEFEF}\textit{Comparison under similar MME scores ($\approx$1680)}}   \\                                                                               
\multicolumn{1}{c|}{FastV~\cite{fastv}}  & \multicolumn{1}{c}{135}                                                        & \multicolumn{1}{c}{189.55}                                     & \multicolumn{1}{c}{1.50$\times$}                          & \multicolumn{1}{c}{3.64}                           & \multicolumn{1}{c}{42.92\%}                              & 1676.49                         \\ 
\multicolumn{1}{c|}{SparseVLM~\cite{zhang2024sparsevlm}}    & \multicolumn{1}{c}{75}                                                    & \multicolumn{1}{c}{176.92}                                     & \multicolumn{1}{c}{1.61$\times$}                          & \multicolumn{1}{c}{5.07}                           & \multicolumn{1}{c}{59.76\%}                              & 1687.45                         \\ 
\multicolumn{1}{c|}{ParVTS (Ours)}            & \multicolumn{1}{c}{\textbf{46}}                                                 & \multicolumn{1}{c}{\textbf{161.33}}                                     & \multicolumn{1}{c}{\textbf{1.77$\times$}}                          & \multicolumn{1}{c}{\textbf{2.57}}                           & \multicolumn{1}{c}{\textbf{30.31\%}}                              & \textbf{1688.63    }                     \\ \bottomrule
\end{tabular}
}
\vspace{-4pt}
\end{table*}

\begin{table*}[]
\caption{
Ablation studies of migration depth $n$ and scheduling strategy on LLaVA-1.5-7B under different vision token budgets.
}
\label{tab:ablation_n}
\renewcommand{\arraystretch}{0.9}
\scriptsize
\centering
\resizebox{0.8\textwidth}{!}{%
\begin{tabular}{c|ccccccc|c}
\toprule
\multicolumn{1}{c|}{\textbf{Method}}                    & \multicolumn{1}{c}{\textbf{GQA}} & \multicolumn{1}{c}{\textbf{MMB}} & \multicolumn{1}{c}{\textbf{MME}} & \multicolumn{1}{c}{\textbf{POPE}} & \multicolumn{1}{c}{\textbf{SQA}} & \multicolumn{1}{c}{\textbf{VizWiz}} & \multicolumn{1}{c|}{\textbf{OCRBench}} & \multicolumn{1}{c}{\textbf{Avg. $\uparrow$}} \\ \hline
\rowcolor[HTML]{EFEFEF} 
\multicolumn{1}{c|}{\cellcolor[HTML]{EFEFEF}LLaVA-1.5-7B~\cite{liu2023llava}} & \multicolumn{7}{c|}{\cellcolor[HTML]{EFEFEF}\textit{All Tokens (576, 100.00\%)}}                                                                                                                        & \textbf{}       \\ \multicolumn{1}{c|}{Vanilla}             & 61.9         & 64.1                             & 1866.1                           & 85.9                              & 69.6                             & 54.3                                & \multicolumn{1}{c|}{313.0 }                                 & 100.00\%  \  \\ \hline

\multicolumn{1}{c|}{\cellcolor[HTML]{EFEFEF}}                                     & \multicolumn{7}{c|}{\cellcolor[HTML]{EFEFEF}\textit{Token Reduction (192, $-$66.67\%)}}             & \textbf{\cellcolor[HTML]{EFEFEF}}       \\

\multicolumn{1}{c|}{$n=3$}     & 59.7          & \textbf{63.6}                     & \textbf{1831.9}                   & \textbf{84.6}                      & 68.5                              & 55.2                                 & \multicolumn{1}{c|}{300.0}                                  & \textbf{97.95\%}                              \\

\multicolumn{1}{c|}{$n=7 $}         & 59.8           & 63.1                              & 1809.0                            & 84.5                               & 68.2                              & 54.93                                & \multicolumn{1}{c|}{299.0 }                                 & 96.96\%      \\

\multicolumn{1}{c|}{$n=11$ }      & 59.9                              & 62.0                              & 1791.8                            & \textbf{84.6}                      & \textbf{68.6}                     & 55.34                                & \multicolumn{1}{c|}{\textbf{303.0} }                        & 96.43\%                                       \\

\multicolumn{1}{c|}{$n=15$}    & \textbf{60.5}       & 61.3              & 1785.7         & 82.8       & \textbf{68.6}                     & \textbf{56.12}          & \multicolumn{1}{c|}{296.0  }                                & 95.86\%                                       \\

\multicolumn{1}{c|}{Non-Sub-FS}    & 59.5                              & 63.2                              & 1825.8                            & 83.6                               & 68.0                              & 53.6                                 & \multicolumn{1}{c|}{293.0 }                                 & 97.28\%                                       \\

\multicolumn{1}{c|}{Sub-FS}    & 57.1                              & 52.7                             & 1531.5                            & 76.7                               & 65.8                              & 53.2                                 & \multicolumn{1}{c|}{119.0 }                                 & 77.77\%                                       \\ \hline

\rowcolor[HTML]{EFEFEF} 
\multicolumn{1}{c|}{\cellcolor[HTML]{EFEFEF}}             & \multicolumn{7}{c|}{\cellcolor[HTML]{EFEFEF}\textit{Token Reduction (128, $-$77.78\%)}}                                                                                                                                                                                       & \textbf{}                                     \\
\multicolumn{1}{c|}{$n=3$}                                                     & 58.2                              & \textbf{63.1}                     & 1789.2                            & 82.0                               & 68.6                              & 55.3                                 & \multicolumn{1}{c|}{297.0}                                  & 95.96\%                                       \\
\multicolumn{1}{c|}{$n=7$}                                                   & 58.3                              & 63.0                              & \textbf{1811.6}                   & 82.2                               & 68.5                              & 55.0                                 & \multicolumn{1}{c|}{\textbf{298.0}}                         & \textbf{96.87\%}                              \\
\multicolumn{1}{c|}{$n=11$}                                                   & 58.6                              & 62.8                              & \textbf{1811.6}                   & \textbf{82.5}                      & 68.9                              & 55.3                                 & \multicolumn{1}{c|}{295.0 }                                 & 96.81\%                                       \\
\multicolumn{1}{c|}{$n=15$  }                                                   & \textbf{60.0}                     & 62.3                              & 1763.0                            & 82.4                               & \textbf{69.1}                             & \textbf{55.9}                        & \multicolumn{1}{c|}{288.0}                                  & 94.65\%                                       \\
\multicolumn{1}{c|}{Non-Sub-FS}                          & 58.2                              & 62.9                              & 1766.2                            & 81.0                               & 68.3                              & 54.1                                 & \multicolumn{1}{c|}{291.0}                                  & 94.69\%                                       \\

\multicolumn{1}{c|}{Sub-FS}    &  58.8                            &     57.6                      &    1651.5                    &        80.3                       &  66.9                      &        53.7                       & \multicolumn{1}{c|}{168.0 }                                 & 84.96\%                                       \\ \hline

\rowcolor[HTML]{EFEFEF} 
\multicolumn{1}{c|}{\cellcolor[HTML]{EFEFEF}}             & \multicolumn{7}{c|}{\cellcolor[HTML]{EFEFEF}\textit{Token Reduction (64, $-$88.89\%)}}                                                                                                                                                                                        & \textbf{}                                     \\
\multicolumn{1}{c|}{$n=3$}                                                    & 55.2                              & 61.3                              & 1739.4                            & 76.9                               & 68.9                              & 55.1                                 & \multicolumn{1}{c|}{\textbf{282.0} }                        & 92.91\%                                       \\
\multicolumn{1}{c|}{$n=7$}                                                    & 55.4                              & 62.3                              & 1722.7                            & 76.8                               & 68.8                              & 54.9                                 & \multicolumn{1}{c|}{278.0 }                                 & 92.19\%                                       \\

\multicolumn{1}{c|}{$n=11$}                                                   & 55.9                              & 61.7                              & 1725.1                            & 77.6                               & 69.0                              & 55.0                                 & \multicolumn{1}{c|}{280.0}                                  & 92.42\%                                       \\

\multicolumn{1}{c|}{$n=15$ }                                                    & \textbf{58.7}                     & \textbf{62.6}                     & \textbf{1773.5}                   & \textbf{80.3}                      & \textbf{69.2}                     & \textbf{55.3}                        &\multicolumn{1}{c|}{ 274.0  }                                & \textbf{94.38\%}                              \\

\multicolumn{1}{c|}{Non-Sub-FS }                         & 55.2                              & 61.2                              & 1671.9                            & 76.0                               & 68.7                              & 54.4                                 & \multicolumn{1}{c|}{271.0 }                                 & 89.80\%    \\ 

\multicolumn{1}{c|}{Sub-FS}    &    54.6                          &    61.7                       &   1636.8                     &      79.3                         &    66.5                    &       53.2                        & \multicolumn{1}{c|}{210.0 }                                 & 85.97\%                                       \\

\bottomrule                                   
\end{tabular}
}
\vspace{-6pt}
\end{table*}

\begin{table*}[!t]
\renewcommand{\arraystretch}{0.9}
\caption{Ablation studies of fusion weights $
\alpha$ and $\beta$ on LLaVA-1.5-7B.}
\label{tab:alpha_beta_ablation_short}
\centering
\resizebox{0.8\textwidth}{!}{%
\begin{tabular}{c|ccccccc|c}
\toprule
\multicolumn{1}{c|}{\textbf{Method}}                    & \multicolumn{1}{c}{\textbf{GQA}}    & \multicolumn{1}{c}{\textbf{MMB}}    & \multicolumn{1}{c}{\textbf{MME}}      & \multicolumn{1}{c}{\textbf{POPE}}   & \multicolumn{1}{c}{\textbf{SQA}}    & \multicolumn{1}{c}{\textbf{VizWiz}} & \multicolumn{1}{c|}{\textbf{OCRBench}} & \multicolumn{1}{c}{\textbf{Avg. $\uparrow$}} \\ \hline
\rowcolor[HTML]{EFEFEF} 
LLaVA-1.5-7B~\cite{liu2023llava}                                             & \multicolumn{7}{c|}{\cellcolor[HTML]{EFEFEF}\textit{All tokens (576,   100.00\%)}}                                                                                                                                                                                                  &                                               \\
Vanilla                                                  & 61.9                                 & 64.1                                 & 1866.1                                 & 85.9                                 & 69.6                                 & 54.3                                 & 313.0                                  & 100.00\%                                      \\ \hline

\rowcolor[HTML]{EFEFEF} 

& \multicolumn{7}{c}{\cellcolor[HTML]{EFEFEF}\textit{Token Reduction (64,   $-$88.89\%)}}                                                                                                                                                                                            &                                               \\
$\alpha$=0.0   $\beta$=1.0                               & 53.1                                 & 58.8                                 & 1723.5                                 & 75.2                                 & 66.3                                 & 52.6                                 & 273.0                                  & 91.55\%                                       \\
$\alpha$=0.1   $\beta$=0.9                               & 52.9                                 & 58.9                                 & 1724.3                                 & 75.3                                 & 66.3                                 & 52.7                                 & 273.0                                  & 91.58\%                                       \\
$\alpha$=0.3   $\beta$=0.7                               & 54.9                                 & 60.6                                 & 1732.9                                 & 76.4                                 & 68.1                                 & 54.9                                 & 276.0                                  & 92.39\%                                       \\
{\color[HTML]{000000} $\alpha$=0.5   $\beta$=0.5 (Ours)} & {\color[HTML]{333333} \textbf{55.2}} & {\color[HTML]{333333} \textbf{61.3}} & {\color[HTML]{333333} \textbf{1739.4}} & {\color[HTML]{333333} \textbf{76.9}} & {\color[HTML]{333333} \textbf{68.9}} & {\color[HTML]{333333} \textbf{55.1}} & {\color[HTML]{333333} \textbf{280.0}}  & {\color[HTML]{333333} \textbf{92.91\%}}       \\
$\alpha$=0.7   $\beta$=0.3                               & 54.8                                 & 60.8                                 & 1733.5                                 & 76.3                                 & 68.7                                 & 54.6                                 & 276.0                                  & 92.43\%                                       \\
$\alpha$=0.9  $\beta$=0.1                                & 55.1                                 & 61.8                                 & 1738.1                                 & 76.6                                 & 68.7                                 & 54.6                                 & 279.0                                  & 92.79\%                                       \\
$\alpha$=1.0  $\beta$=0.0                                & 55.1                                 & 60.8                                 & 1734.7                                 & 76.5                                 & 68.4                                 & 54.6                                 & 277.0                                  & 92.52\%                          \\ \bottomrule        
\end{tabular}
}
\vspace{-10pt}
\end{table*}

\vspace{-2pt}
\subsection{Cost and Efficiency Analysis}

We compare different methods under three configurations with similar MME~\cite{mme} scores, reporting the number of retained visual tokens, inference latency, and TFLOPs cost at each accuracy level. As shown in Table\,\ref{tab:latency}, ParVTS consistently achieves the highest inference accuracy with fewer tokens and lower computational cost. 
To further evaluate efficiency, we provide fine-grained empirical statistics and analysis across diverse concurrency and response-length settings, reporting latency, GPU peak memory, and TFLOPs for both the prefilling and decoding stages in Appendix\,\ref{sec:detail_latency}.

Moreover, we establish a theoretical speedup model, which provides an analysis of how the pruning rate and migration depth influence the acceleration of both the prefilling and decoding stages, as detailed in Appendix\,\ref{sec:theoretical}}.

ParVTS also offers better engineering compatibility for efficient deployment. Unlike FastV~\cite{fastv}, PruMerge~\cite{prumerge}, and HiRED~\cite{arif2024hired}, which require access to intermediate attention matrices and thus conflict with the computation pattern of Flash-Attention~\cite{dao2022flashattention,dao2023flashattention2}, ParVTS remains fully compatible with both Flash-Attention and KV-cache reuse, ensuring seamless integration in real-world deployment.

\vspace{-2pt}
\subsection{Ablation Studies}

\noindent \textbf{Migration Depth $n$.}
We first investigate how the migration depth $n$ affects model performance. As shown in Table\,\ref{tab:ablation_n}, a larger migration depth $n$ is required to maintain performance under more aggressive pruning. 

This trend supports the visual information migration principle: under tighter budgets, more non-subject tokens are pruned after layer $n$, requiring a larger $n$ to ensure sufficient semantic migration before they are discarded. In practice, we set $n=3$ for LLaVA-1.5-7B and $n=16$ for LLaVA-Next-7B as default values, to balance accuracy and efficiency. Higher values of $n$ may further improve performance when resources permit.

\noindent \textbf{Scheduling Strategy.}
We further compare two sequential scheduling variants: Subject-First Scheduling (Sub-FS) and Non-Subject-First Scheduling (Non-Sub-FS). As shown in Table\,\ref{tab:ablation_n}, both perform consistently worse than our parallel strategy, under the same migration depth $n=3$, highlighting the structural limitations of sequential token usage. More detailed results with different switching depths are provided in Appendix\,\ref{appendix:scheduling_variants}.

\noindent \textbf{Fusion Weights $\alpha$ and $\beta$.}
We ablate the fusion weights $\alpha$ and $\beta$ of Eq.\,(\ref{eq:alpha}) in Table~\ref{tab:alpha_beta_ablation_short}. The results show that setting $\alpha=0.5$ and $\beta=0.5$ consistently achieves the best performance across benchmarks. More detailed results under different pruning rates are provided in Appendix\,\ref{appendix:alpha_beta_ablation}.

\noindent \textbf{Reliability of [CLS]-Based Subject Localization.}
We provide visualization results in Appendix\,\ref{sec:visual_cls_token}, showing that the [CLS] token attention effectively separates subject and non-subject regions. This confirms the reliability of using [CLS]-based saliency for foreground localization.

\section{Limitations and Future Work}
\label{sec:limitations}
ParVTS uses attention to the [CLS] token in the vision encoder to separate subject from non-subject tokens. While this lightweight, supervision-free strategy aligns with our training-free design, it may struggle in images with multiple salient regions, subtle foregrounds, or non-object-centric queries. Future work could explore more robust, adaptive token grouping methods to enhance visual information selection.
Besides, key hyperparameters (the migration depth $n$ and fusion weights $\alpha$, $\beta$) are empirically set. Auto-tuning them by input or task remains an open direction.

\section{Conclusion}
We present ParVTS, a training-free vision token scheduling framework that leverages early-layer information migration and parallel execution to recover non-subject semantics, enabling fast and accurate MLLM inference.  
Experiments across diverse benchmarks and compression levels show that ParVTS consistently achieves strong performance while substantially reducing inference cost.  
These results highlight the potential of leveraging intrinsic model behaviors for efficient inference, offering new insights into mechanism-aware multimodal reasoning.

\bibliographystyle{unsrt}  
\bibliography{references}

\clearpage
\setcounter{page}{1}

\section*{\centering \LARGE Appendix}
\appendix

\section{Subject-Question Statistics in Figure\,\ref{fig:percentage}}
\label{sec:subject_annotation}

To compute the subject-oriented question proportions shown in Figure\,\ref{fig:percentage}, we use LLaVA-1.5-7B~\cite{liu2023llava} to classify whether each question in four VQA datasets (SQA~\cite{lu2022sqa}, AI2D~\cite{ai2d}, OCRBench~\cite{liu2024ocrbench}, TextVQA~\cite{Singh_2019_CVPR_textvqa}) refers to the main subject in the image. The prompt used is:

\begin{quote}
\small
\ttfamily
The main subject in the image refers to the most visually important object, such as a person or object that is the focus of the image. Even if the question only asks about a part or property of that object, it still counts as referring to the main subject. \\
Does the question "\{question\}" ask about the main subject?
\end{quote}

LLaVA's binary responses ("Yes"/"No") are used to label each question, and the subject-oriented question percentage is then computed per dataset accordingly.

\section{Implementation Details\label{sec:implementation}}

In our implementation, the parallel scheduling is activated from the second transformer layer onward; for simplicity, the first layer---where all tokens are processed jointly---is omitted in both Figure\,\ref{fig:framework} and Sec.\,\ref{sec:parallel}. The fusion weights $\alpha$ and $\beta$ in Sec.\,\ref{sec:parallel} are set to 0.5.
All experiments, except for Video-LLaVA, are conducted using the lmms-eval~\cite{li2024lmmseval} evaluation framework.

The migration depth $n$ used in our experiments is summarized in Table\,\ref{tab:appendix_n}. Models with a larger number of visual tokens typically require deeper transformer layers to complete effective information migration, ensuring that both subject and non-subject semantics are sufficiently integrated into the question tokens before pruning.

\begin{table}[h]
\renewcommand{\arraystretch}{1.6}
\caption{Migration depth $n$ settings across different MLLM backbones.}
\label{tab:appendix_n}
\footnotesize
\centering
\resizebox{0.5\textwidth}{!}{%
\begin{tabular}{c|c|c|c}
\toprule
\textbf{Backbone}               & \textbf{\#Params} & \textbf{Vision Token Numbers} & \textbf{Migration Depth $n$} \\ \hline
\multirow{2}{*}{LLaVA-1.5}   & 7B             & 576                 & 3                            \\ 
                             & 13B            & 576                 & 3                            \\ \hline
\multirow{2}{*}{LLaVA-Next}  & 7B             & 2880                & 16                           \\
                             & 13B            & 2880                & 16                           \\ \hline
\multirow{2}{*}{Qwen2.5-VL}  & 3B             & Dynamic             & 18                           \\
                             & 7B             & Dynamic             & 18                           \\ \hline
\multirow{3}{*}{Qwen3-VL}  & 2B             & Dynamic             & 10                           \\
& 4B             & Dynamic             & 12                           \\
                             & 8B             & Dynamic             & 12                           \\ \hline

\multirow{2}{*}{InternVL2}   & 2B             & Dynamic             & 18                           \\
                             & 8B             & Dynamic             & 16                           \\ \hline
                             
\multirow{2}{*}{InternVL2.5} & 2B             & Dynamic             & 18                           \\
                             & 8B             & Dynamic             & 16                           \\ \hline

Video-LLaVA                  & 7B             & 2056                & 24           \\ \bottomrule               
\end{tabular}
}
\end{table}
\vspace{-4pt}

\section{Experiments on More Token Budgets and MLLM Backbones\label{sec:more}}

Table\,\ref{tab:llava1.5-7b-full} \textasciitilde Table\,\ref{tab:internvl25-8b} provide additional experimental results covering a broader range of multimodal models, including LLaVA-1.5~\cite{liu2023llava}, LLaVA-Next~\cite{liu2024improvedllava}, Qwen2.5-VL~\cite{bai2025qwen2}, Qwen3-VL, InternVL2~\cite{chen2024internvl15}, InternVL2.5~\cite{chen2024internvl25}. The results span different model sizes (e.g., 2B, 7B, 13B) and various vision token reduction ratios to validate the generality of ParVTS across architectures and settings.
Across all model backbones and token budgets, ParVTS demonstrates solid and consistent performance, further indicating its scalability across architectures.

Notably, on LLaVA-Next-13B, ParVTS achieves impressively high average scores, even surpassing the all-token baseline. This indicates that early-stage migration from non-subject tokens helps maintain performance under reduction, and the improvement over the baseline may be attributed to the removal of noisy or redundant visual tokens, which allows the model to focus more effectively on salient regions during inference.

For Qwen2.5-VL, ParVTS outperforms FastV~\cite{fastv} across all reduction ratios on both 3B and 7B variants, especially under higher reduction rates (e.g., 77.78\% and 88.89\%), where it shows clear advantages on datasets such as MME~\cite{mme}, POPE~\cite{Li-hallucination-2023}, and OCRBench~\cite{liu2024ocrbench}.

On Qwen3-VL, InternVL2.5 and InternVL2, ParVTS consistently improves over FastV under dynamic reduction settings, demonstrating its scalability across model sizes.

Moreover, we observe that as the token reduction ratio increases, the performance gap between ParVTS and other methods becomes more noticeable. This trend clearly suggests ParVTS's relative advantage when operating under stricter token constraints.

\begin{table*}[!b]
\renewcommand{\arraystretch}{1.2}
\caption{
Comparative experiments on \textbf{LLaVA-1.5-7B} under different vision token budgets. \textbf{Bold} indicates the best result.}
\label{tab:llava1.5-7b-full}
\resizebox{\textwidth}{!}{%
\begin{tabular}{c|cccccccc|c}
\toprule
\multicolumn{1}{c|}{\textbf{Methods}}         & \multicolumn{1}{c}{\textbf{GQA}}           & \multicolumn{1}{c}{\textbf{MMB}}           & \multicolumn{1}{c}{\textbf{MME}}             & \multicolumn{1}{c}{\textbf{POPE}}          & \multicolumn{1}{c}{\textbf{SQA}}           & \multicolumn{1}{c}{\textbf{VQA\textsuperscript{v2}}}                      & \multicolumn{1}{c}{\textbf{VizWiz}}        & \multicolumn{1}{c|}{\textbf{OCRBench}}       & \textbf{Avg.$\uparrow$}                       \\ \hline

\rowcolor[HTML]{EFEFEF} 
\multicolumn{1}{c|}{\cellcolor[HTML]{EFEFEF}LLaVA-1.5-7B~\cite{liu2023llava}} & \multicolumn{8}{c|}{\cellcolor[HTML]{EFEFEF}\textit{All Tokens (576, 100.00\%)}}    & \multicolumn{1}{c}{\cellcolor[HTML]{EFEFEF}}     \\ 

\multicolumn{1}{c|}{Vanilla}    & \multicolumn{1}{c}{61.9}         & \multicolumn{1}{c}{64.1}         & \multicolumn{1}{c}{1866.1}         & \multicolumn{1}{c}{85.9}         & \multicolumn{1}{c}{69.6}         & \multicolumn{1}{c}{78.5}                      & \multicolumn{1}{c}{54.3}         & \multicolumn{1}{c|}{313.0}         & 100.00 \%                       \\ \hline

  \rowcolor[HTML]{EFEFEF} 
\multicolumn{1}{c|}{\cellcolor[HTML]{EFEFEF}} & \multicolumn{8}{c|}{\cellcolor[HTML]{EFEFEF}\textit{Token Reduction (192, $-$66.67\%)}}  & \multicolumn{1}{c}{\cellcolor[HTML]{EFEFEF}}                                                  \\ 

\multicolumn{1}{c|}{ToMe~\cite{bolya2022tome}}           & \multicolumn{1}{c}{54.3}          & \multicolumn{1}{c}{60.5}          & \multicolumn{1}{c}{1563.0}          & \multicolumn{1}{c}{72.4}          & \multicolumn{1}{c}{65.2}          & \multicolumn{1}{c}{68.0}                       & \multicolumn{1}{c}{-}             & \multicolumn{1}{c|}{-}              & 96.35\%                    \\
\bottomrule
\end{tabular}
}
\end{table*}

\begin{table*}[!]
\renewcommand{\arraystretch}{1.1}
\caption{
Comparative experiments on \textbf{LLaVA-1.5-13B} under different vision token budgets. \textbf{Bold} indicates the best result.}
\label{tab:llava-1.5-13b}
\scriptsize
\centering
\resizebox{\textwidth}{!}{%
\begin{tabular}{c|cccccc|c}
\toprule
\textbf{Method} & \multicolumn{1}{c}{\textbf{GQA}} & \multicolumn{1}{c}{\textbf{MMB}} & \multicolumn{1}{c}{\textbf{MME}} & \multicolumn{1}{c}{\textbf{POPE}} & \multicolumn{1}{c}{\textbf{SQA}} & \textbf{VizWiz} & \textbf{Avg. $\uparrow$} \\ \hline
\rowcolor[HTML]{EFEFEF} 
LLaVA-1.5-13B~\cite{liu2023llava}   & \multicolumn{6}{c|}{\cellcolor[HTML]{EFEFEF}\textit{All Tokens (576, 100.00\%)}}                                                                                                                             &                          \\  
Vanilla         & 63.3                              & 68.6                              & 1825.5                            & 87.1                               & 72.8                              & 56.6            & 100.00\%                 \\  
\hline
\rowcolor[HTML]{EFEFEF} 
                & \multicolumn{6}{c|}{\cellcolor[HTML]{EFEFEF}\textit{Token Reduction (432, $-$25.00\%)}}                                                                                                                           &                          \\  
FastV~\cite{fastv}           & 63.1                              & \textbf{68.4}                     & 1544.0                            & 85.9                               & 72.8                              & 53.8            & 86.85\%                  \\  
SparseVLM~\cite{zhang2024sparsevlm}       & 61.3                              & 68.3                              & 1481.4                            & 72.5                               & \textbf{73.7}                     & 53.6            & 83.30\%                  \\  
FasterVLM~\cite{zhang2024fastervlm}       & 61.2                              & 67.6                              & 1487.3                            & 86.7                               & 73.0                              & 53.2            & 84.14\%                  \\
\bottomrule
\end{tabular}
}
\end{table*}

\begin{table*}[!]
\renewcommand{\arraystretch}{1.0}
\caption{Comparative experiments on \textbf{LLaVA-Next-7B} under different vision token budgets. \textbf{Bold} indicates the best result.}
\label{tab:llavanext-7b}
\scriptsize
\centering
\resizebox{0.85\textwidth}{!}{%
\begin{tabular}{c|cccccc|c}
\toprule
\textbf{Method}                               & \textbf{GQA}  & \textbf{MMB}  & \textbf{MME}    & \textbf{POPE} & \textbf{SQA}  & \textbf{VizWiz} & \textbf{Avg. $\uparrow$} \\ \hline
\rowcolor[HTML]{EFEFEF} 
LLaVA-Next-7B~\cite{liu2024improvedllava}                                 & \multicolumn{6}{c|}{\cellcolor[HTML]{EFEFEF}\textit{All Tokens (2880, 100.00\%)}}                  &                          \\
Vanilla                                       & 62.5          & 66.2          & 1875.1          & 87.8          & 68.0          & 59.7            & 100.00\%                 \\ \hline
\rowcolor[HTML]{EFEFEF} 
                                              & \multicolumn{6}{c|}{\cellcolor[HTML]{EFEFEF}\textit{Token Reduction (2160,   $-$25.00\%)}}         &                          \\
FastV~\cite{fastv}                                         & 62.5          & 67.6          & 1506.3          & 86.3          & 69.3          & 55.1            & 83.23\%                  \\
SparseVLM~\cite{zhang2024sparsevlm}                                     & 62.6          & 66.1          & 1507.8          & 73.2          & 68.5          & 55.2            & 82.61\%                  \\
FasterVLM~\cite{zhang2024fastervlm}                                     & 63.7          & 67.5          & 1533.4          & 87.5          & 68.4          & 56.5            & 84.57\%                  \\
GlobalCom\textsuperscript{2}~\cite{liu2025globalcom} & \textbf{63.8} & \textbf{68.0} & 1548.4          & \textbf{87.8} & \textbf{68.7} & 56.5            & 85.30\%                  \\
\bottomrule     
\end{tabular}
}
\end{table*}

\begin{table*}[!]
\renewcommand{\arraystretch}{1.0}
\caption{
Comparative experiments on \textbf{LLaVA-Next-13B} under different token budgets. \textbf{Bold} indicates the best result.}
\label{tab:llavanext-13b}
\scriptsize
\centering
\resizebox{0.85\textwidth}{!}{%
\begin{tabular}{c|cccccc|c}
\toprule
\textbf{Method} & \multicolumn{1}{c}{\textbf{GQA}} & \multicolumn{1}{c}{\textbf{MMB}} & \multicolumn{1}{c}{\textbf{MME}} & \multicolumn{1}{c}{\textbf{POPE}} & \multicolumn{1}{c}{\textbf{SQA}} & \textbf{VizWiz} & \textbf{Avg. $\uparrow$} \\ \hline
\rowcolor[HTML]{EFEFEF} 
LLaVA-Next-13B~\cite{liu2024improvedllava}  & \multicolumn{6}{c|}{\cellcolor[HTML]{EFEFEF}\textit{All Tokens (2880, 100.00\%)}}                                                                                                                            &                          \\  
Vanilla         & 65.4                              & 70.0                              & 1575.9                            & 86.2                               & 73.5                              & 60.5            & 100.00\%                 \\  \hline
\rowcolor[HTML]{EFEFEF} 
                & \multicolumn{6}{c|}{\cellcolor[HTML]{EFEFEF}\textit{Token Reduction (2160, $-$25.00\%)}}                                                                                                                          &                          \\  
FasterVLM~\cite{zhang2024fastervlm}       & 64.6                              & \textbf{69.8}                     & 1560.3                            & 87.6                               & 72.6                              & 58.7            & 99.08\%                  \\  
GlobalCom\textsuperscript{2}~\cite{liu2025globalcom}      & \textbf{65.0}                     & 69.2                              & 1567.9                            & \textbf{87.8}                      & \textbf{72.8}                     & 58.6            & 99.48\%                  \\  
ParVTS (Ours)            & 63.4                              & 64.1                              & \textbf{1846.0}                   & 84.6                               & 71.1                              & \textbf{61.3}   & \textbf{113.42\%}        \\
\bottomrule
\end{tabular}
}
\end{table*}

\begin{table*}[!]
\renewcommand{\arraystretch}{1.2}
\caption{Comparative experiments on \textbf{Qwen2.5-VL-3B} under different token budgets. \textbf{Bold} indicates the best result.}
\label{tab:qwen-3b}
\scriptsize
\centering
\resizebox{0.9\textwidth}{!}{%
\begin{tabular}{c|ccccccc|c}
\toprule
\textbf{Method} & \textbf{GQA}  & \textbf{MMB}  & \textbf{MME}    & \textbf{POPE} & \textbf{SQA}  & \textbf{OCRBench} & \textbf{VizWiz} & \textbf{Avg. $\uparrow$} \\  \hline
\rowcolor[HTML]{EFEFEF} 
Qwen2.5-VL-3B   & \multicolumn{7}{c|}{\cellcolor[HTML]{EFEFEF}\textit{All tokens (Dynamic, 100.00\%)}}                                   &                          \\
Vanilla         & 60.0          & 78.6          & 2181.7          & 87.7          & 50.6          & 775.0             & 68.5            & 100.00\%                 \\ \hline
\rowcolor[HTML]{EFEFEF} 
                & \multicolumn{7}{c|}{\cellcolor[HTML]{EFEFEF}\textit{Dynamic Token Reduction ($-$66.67\%)}}                            &                          \\ 
FastV~\cite{fastv}           & 55.5          & 75.0          & 2005.8          & 86.0          & 49.6          & \textbf{567.0}    & \textbf{67.4}   & 88.02\%                  \\
ParVTS (Ours)   & \textbf{57.9} & \textbf{75.6} & \textbf{2033.0} & \textbf{86.3} & \textbf{67.2} & 547.0             & 67.0            & \textbf{88.86\%}         \\  \hline
\rowcolor[HTML]{EFEFEF} 
                & \multicolumn{7}{c|}{\cellcolor[HTML]{EFEFEF}\textit{Dynamic Token Reduction ($-$77.78\%)}}                            &                          \\
FastV~\cite{fastv}           & 52.6          & 71.6          & 1904.1          & 82.7          & 48.6          & 442.0             & \textbf{66.3}   & 80.80\%                  \\
\bottomrule
\end{tabular}
}
\end{table*}

\begin{table*}[!]
\renewcommand{\arraystretch}{1.2}
\caption{Comparative experiments on \textbf{Qwen2.5-VL-7B} under different token budgets. \textbf{Bold} indicates the best result.}
\label{tab:qwen-7b}
\scriptsize
\centering
\resizebox{0.9\textwidth}{!}{%
\begin{tabular}{c|ccccccc|c}
\toprule
\textbf{Method} & \textbf{GQA}  & \textbf{MMB}  & \textbf{MME}    & \textbf{POPE} & \textbf{SQA}  & \textbf{OCRBench} & \textbf{VizWiz} & \textbf{Avg. $\uparrow$} \\ \hline
\rowcolor[HTML]{EFEFEF} 
Qwen2.5-VL-7B   & \multicolumn{7}{c|}{\cellcolor[HTML]{EFEFEF}\textit{All tokens (Dynamic, 100.00\%)}}                                   &                          \\
Vanilla         & 60.4          & 84.1          & 2321.5          & 87.7          & 77.1          & 822.0             & 70.8            & 100.00\%                 \\ \hline
\rowcolor[HTML]{EFEFEF} 
                & \multicolumn{7}{c|}{\cellcolor[HTML]{EFEFEF}\textit{Dynamic Token Reduction ($-$66.67\%)}}                            &                          \\
FastV~\cite{fastv}            & 55.6          & 79.4          & 2230.5          & 85.3          & \textbf{76.2} & 551.0             & \textbf{69.9}   & 89.33\%                  \\
ParVTS (Ours)   & \textbf{59.0} & \textbf{79.6} & \textbf{2255.5} & \textbf{85.7} & 73.2          & \textbf{597.0}    & 67.3            & \textbf{91.30\%}         \\ \hline
\rowcolor[HTML]{EFEFEF} 
                & \multicolumn{7}{c|}{\cellcolor[HTML]{EFEFEF}\textit{Dynamic Token Reduction ($-$77.78\%)}}                            &                          \\
FastV~\cite{fastv}            & 52.6          & 75.9          & 2111.6          & 82.1          & \textbf{74.2} & 408.0             & \textbf{68.6}   & 81.53\%                  \\
\bottomrule
\end{tabular}
}
\end{table*}

\begin{table*}[!]
\renewcommand{\arraystretch}{1.2}
\caption{Comparative experiments on \textbf{Qwen3-VL-2B} under different token budgets. \textbf{Bold} indicates the best result.}
\label{tab:qwen3-2b}
\scriptsize
\centering
\resizebox{0.9\textwidth}{!}{%
\begin{tabular}{c|ccccccc|c}
\toprule
\textbf{Method} & \textbf{GQA}  & \textbf{MMB}  & \textbf{MME}    & \textbf{POPE} & \textbf{SQA}  & \textbf{OCRBench} & \textbf{VizWiz} & \textbf{Avg. $\uparrow$} \\ \hline
\rowcolor[HTML]{EFEFEF} 
Qwen3-VL-2B   & \multicolumn{7}{c|}{\cellcolor[HTML]{EFEFEF}\textit{All tokens (Dynamic, 100.00\%)}}                                   &                          \\
Vanilla         & 59.4            & 76.0            & 2009.0            & 90.0            & 86.5            & 806.0              & 67.7            & 100.00\%                 \\ \hline
\rowcolor[HTML]{EFEFEF} 
                & \multicolumn{7}{c|}{\cellcolor[HTML]{EFEFEF}\textit{Dynamic Token Reduction ($-$66.67\%)}}                            &                          \\
FastV~\cite{fastv}            & 52.1            & 73.5            & 1891.5            & 88.0            & 83.3            & 575.0              & 67.0            & 88.60\%                  \\
ParVTS (Ours)   & \textbf{59.3}   & \textbf{75.9}   & \textbf{2001.3}   & \textbf{89.9}   & \textbf{86.0}   & \textbf{640.0}     & \textbf{67.1}   & \textbf{94.52\%}         \\ \hline
\rowcolor[HTML]{EFEFEF} 
                & \multicolumn{7}{c|}{\cellcolor[HTML]{EFEFEF}\textit{Dynamic Token Reduction ($-$77.78\%)}}                            &                          \\
FastV~\cite{fastv}                          & 52.1            & 73.5            & 1891.5            & 88.0            & 83.3            & 575.0              & 67.0            & 88.60\%                  \\
\bottomrule
\end{tabular}
}
\end{table*}

\begin{table*}[!]
\renewcommand{\arraystretch}{1.2}
\caption{Comparative experiments on \textbf{Qwen3-VL-4B} under different token budgets. \textbf{Bold} indicates the best result.}
\label{tab:qwen3-4b}
\scriptsize
\centering
\resizebox{0.9\textwidth}{!}{%
\begin{tabular}{c|ccccccc|c}
\toprule
\textbf{Method} & \textbf{GQA}  & \textbf{MMB}  & \textbf{MME}    & \textbf{POPE} & \textbf{SQA}  & \textbf{OCRBench} & \textbf{VizWiz} & \textbf{Avg. $\uparrow$} \\ \hline
\rowcolor[HTML]{EFEFEF} 
Qwen3-VL-4B   & \multicolumn{7}{c|}{\cellcolor[HTML]{EFEFEF}\textit{All tokens (Dynamic, 100.00\%)}}                                   &                          \\
Vanilla                                                  & 62.6            & 84.4            & 2334.5            & 89.6            & 93.3            & 853.0              & 69.3            & 100.00\%                 \\ \hline

\rowcolor[HTML]{EFEFEF} 
                & \multicolumn{7}{c|}{\cellcolor[HTML]{EFEFEF}\textit{Dynamic Token Reduction ($-$66.67\%)}}                            &                          \\
FastV~\cite{fastv}                         & 56.0            & 77.3            & 1962.0            & 85.1            & 83.3            & 636.0              & 66.6            & 82.71\%                  \\

ParVTS (Ours)                                          & \textbf{62.1}   & \textbf{83.8}   & \textbf{2299.9}   & \textbf{89.4}   & \textbf{91.8}   & \textbf{674.0}     & \textbf{68.8}   & \textbf{93.96\%}         \\ \hline

\rowcolor[HTML]{EFEFEF} 
                & \multicolumn{7}{c|}{\cellcolor[HTML]{EFEFEF}\textit{Dynamic Token Reduction ($-$77.78\%)}}                            &                          \\
FastV~\cite{fastv}                                      & 51.3            & 71.7            & 1891.1            & 81.8            & 79.2            & 604.0              & 64.0            & 79.27\%                  \\
\bottomrule
\end{tabular}
}
\end{table*}

\begin{table*}[!]
\renewcommand{\arraystretch}{1.2}
\caption{Comparative experiments on \textbf{Qwen3-VL-8B} under different token budgets. \textbf{Bold} indicates the best result.}
\label{tab:qwen3-8b}
\scriptsize
\centering
\resizebox{0.9\textwidth}{!}{%
\begin{tabular}{c|ccccccc|c}
\toprule
\textbf{Method} & \textbf{GQA}  & \textbf{MMB}  & \textbf{MME}    & \textbf{POPE} & \textbf{SQA}  & \textbf{OCRBench} & \textbf{VizWiz} & \textbf{Avg. $\uparrow$} \\ \hline
\rowcolor[HTML]{EFEFEF} 
Qwen3-VL-8B   & \multicolumn{7}{c|}{\cellcolor[HTML]{EFEFEF}\textit{All tokens (Dynamic, 100.00\%)}}                                   &                          \\
Vanilla                                                  & 61.7            & 84.6            & 2384.1            & 89.0            & 94.2            & 859.0              & 69.7            & 100.00\%                 \\ \hline

\rowcolor[HTML]{EFEFEF} 
                & \multicolumn{7}{c|}{\cellcolor[HTML]{EFEFEF}\textit{Dynamic Token Reduction ($-$66.67\%)}}                            &                          \\
FastV~\cite{fastv}                  & 55.5            & 78.4            & 1997.5            & 85.1            & 86.0            & 685.0              & 67.2            & 83.86\%                  \\

ParVTS (Ours)                                            & \textbf{61.4}   & \textbf{83.2}   & \textbf{2374.6}   & \textbf{88.8}   & \textbf{94.0}   & \textbf{691.0}     & \textbf{70.0}   & \textbf{95.07\%}         \\ \hline

\rowcolor[HTML]{EFEFEF} 
                & \multicolumn{7}{c|}{\cellcolor[HTML]{EFEFEF}\textit{Dynamic Token Reduction ($-$77.78\%)}}                            &                          \\
FastV~\cite{fastv}                                          & 52.1            & 74.5            & 1944.1            & 81.9            & 83.2            & 624.0              & 65.7            & 80.32\%                  \\
\bottomrule
\end{tabular}
}
\end{table*}

\begin{table*}[!]
\renewcommand{\arraystretch}{1.0}
\caption{Comparative experiments on \textbf{InternVL2-2B} under different token budgets. \textbf{Bold} indicates the best result.}
\label{tab:internvl2-2b}
\scriptsize
\centering
\resizebox{0.85\textwidth}{!}{%
\begin{tabular}{c|cccccc|c}
\toprule
\textbf{Method} & \textbf{GQA}  & \textbf{MMB}  & \textbf{MME}    & \textbf{POPE} & \textbf{SQA}  & \textbf{AI2D} & \textbf{Avg. $\uparrow$} \\ \hline

\rowcolor[HTML]{EFEFEF} 
InternVL2-2B    & \multicolumn{6}{c|}{\cellcolor[HTML]{EFEFEF}\textit{All tokens (Dynamic, 100.00\%)}}                & \textit{}                \\
Vanilla         & 59.6          & 93.6          & 1819.4          & 88.6          & 71.6          & 72.0          & 100.00\%                 \\ \hline
\rowcolor[HTML]{EFEFEF} 
                & \multicolumn{6}{c|}{\cellcolor[HTML]{EFEFEF}\textit{Dynamic Token Reduction (  $-$77.78\%)}}    &                          \\
FastV~\cite{fastv}           & \textbf{53.1} & 68.0          & 1664.0          & \textbf{83.0} & 89.8          & 67.4          & 91.85\%                  \\
ParVTS (Ours)   & 51.9          & \textbf{69.2} & \textbf{1718.7} & 77.6          & \textbf{90.3} & \textbf{68.7} & \textbf{94.17\%}         \\ \hline
\rowcolor[HTML]{EFEFEF} 
                & \multicolumn{6}{c|}{\cellcolor[HTML]{EFEFEF}\textit{Dynamic Token Reduction (  $-$88.89\%)}}    &                          \\
FastV~\cite{fastv}           & 46.4          & 86.2          & 1557.0          & 72.5          & 60.6          & 63.8          & 85.55\%                  \\
ParVTS (Ours)   & \textbf{48.6} & \textbf{90.7} & \textbf{1612.8} & \textbf{72.8} & \textbf{68.2} & \textbf{68.4} & \textbf{88.96\%}    \\ \bottomrule    
\end{tabular}
}
\end{table*}

\begin{table*}[!]
\renewcommand{\arraystretch}{1.0}
\caption{Comparative experiments on \textbf{InternVL2-8B} under different token budgets. \textbf{Bold} indicates the best result.}
\label{tab:internvl2-8b}
\scriptsize
\centering
\resizebox{0.85\textwidth}{!}{%
\begin{tabular}{c|cccccc|c}
\toprule
\textbf{Method} & \textbf{GQA}  & \textbf{MMB}  & \textbf{MME}    & \textbf{POPE} & \textbf{SQA}  & \textbf{AI2D} & \textbf{Avg. $\uparrow$} \\ \hline
\rowcolor[HTML]{EFEFEF} 
InternVL2-8B    & \multicolumn{6}{c|}{\cellcolor[HTML]{EFEFEF}\textit{All tokens (Dynamic, 100.00\%)}}                & \textit{}                \\
Vanilla         & 62.7          & 96.9          & 2207.8          & 87.9          & 82.0           & 82.4          & 100.00\%                 \\ \hline
\rowcolor[HTML]{EFEFEF} 
                & \multicolumn{6}{c|}{\cellcolor[HTML]{EFEFEF}\textit{Dynamic Token Reduction (  $-$77.78\%)}}    &                          \\
FastV~\cite{fastv}           & \textbf{59.5} & 75.3          & 2110.9          & \textbf{85.7} & 94.0          & 74.4          & 95.43\%                  \\
ParVTS (Ours)   & 51.5          & \textbf{76.7} & \textbf{2162.5} & 80.3          & \textbf{94.7} & \textbf{78.7} & \textbf{97.13\%}         \\ \hline
\rowcolor[HTML]{EFEFEF} 
                & \multicolumn{6}{c|}{\cellcolor[HTML]{EFEFEF}\textit{Dynamic Token Reduction (  $-$88.89\%)}}    &                          \\
FastV~\cite{fastv}           & \textbf{55.0} & 85.6          & 1959.1          & \textbf{79.7} & 67.3          & 71.2          & 88.48\%                  \\
ParVTS (Ours)   & 46.9          & \textbf{95.8} & \textbf{2148.1} & 78.6          & \textbf{74.2} & \textbf{79.0} & \textbf{96.30\%}       \\ \bottomrule 
\end{tabular}
}
\end{table*}

\begin{table*}[!]
\renewcommand{\arraystretch}{1.0}
\caption{Comparative experiments on \textbf{InternVL2.5-2B} under different token budgets. \textbf{Bold} indicates the best result.}
\label{tab:internvl_25_2b}
\scriptsize
\centering
\resizebox{0.85\textwidth}{!}{%
\begin{tabular}{c|cccccc|c}
\toprule
\textbf{Method} & \textbf{GQA}  & \textbf{MMB}  & \textbf{MME}    & \textbf{POPE} & \textbf{SQA}  & \textbf{AI2D}  & \textbf{Avg. $\uparrow$} \\ \hline
\rowcolor[HTML]{EFEFEF} 
InternVL2.5-2B & \multicolumn{6}{c|}{\cellcolor[HTML]{EFEFEF}\textit{All tokens (Dynamic, 100.00\%)}}                 & \textit{}                \\
Vanilla         & 59.5          & 74.2          & 2097.9          & 90.7          & 96.1          & 73.87          & 100.00\%                 \\ \hline
\rowcolor[HTML]{EFEFEF} 
                & \multicolumn{6}{c|}{\cellcolor[HTML]{EFEFEF}\textit{Dynamic Token Reduction (  $-$77.78\%)}}     &                          \\
FastV~\cite{fastv}           & \textbf{54.2} & 70.9          & 2007.3          & \textbf{88.1} & \textbf{90.0} & 67.45          & 95.40\%                  \\
ParVTS (Ours)   & 47.1          & \textbf{71.2} & \textbf{2022.5} & 86.6          & 89.9          & \textbf{67.91} & \textbf{95.71\%}         \\ \hline
\rowcolor[HTML]{EFEFEF} 
                & \multicolumn{6}{c|}{\cellcolor[HTML]{EFEFEF}\textit{Dynamic Token Reduction (  $-$88.89\%)}}     &                          \\
FastV~\cite{fastv}           & \textbf{49.9} & 64.1          & 1931.3          & 84.4          & 85.1          & 63.99          & 91.43\%                  \\
ParVTS (Ours)   & 44.2          & \textbf{70.7} & \textbf{2027.9} & \textbf{85.9} & \textbf{93.2} & \textbf{67.81} & \textbf{95.88\%}     \\ \bottomrule   
\end{tabular}
}
\end{table*}

\begin{table*}[!]
\renewcommand{\arraystretch}{1.0}
\caption{Comparative experiments on \textbf{InternVL2.5-8B} under different token budgets. \textbf{Bold} indicates the best result.}
\label{tab:internvl25-8b}
\scriptsize
\centering
\resizebox{0.85\textwidth}{!}{%
\begin{tabular}{c|cccccc|c}
\toprule
\textbf{Method} & \textbf{GQA}  & \textbf{MMB}  & \textbf{MME}    & \textbf{POPE} & \textbf{SQA}  & \textbf{AI2D} & \textbf{Avg. $\uparrow$}                              \\ \hline
\rowcolor[HTML]{EFEFEF} 
InternVL2.5-8B & \multicolumn{6}{c|}{\cellcolor[HTML]{EFEFEF}\textit{All tokens (Dynamic, 100.00\%)}}                & \multicolumn{1}{l}{\cellcolor[HTML]{EFEFEF}\textit{}} \\
Vanilla         & 63.0          & 84.9          & 2304.8          & 90.6          & 97.9          & 83.0          & 100.00\%                                              \\ \hline
\rowcolor[HTML]{EFEFEF} 
                & \multicolumn{6}{c|}{\cellcolor[HTML]{EFEFEF}\textit{Dynamic Token Reduction (  $-$77.78\%)}}    &                                                       \\
FastV~\cite{fastv}           & \textbf{60.2} & 78.7          & 2194.2          & \textbf{90.4} & 93.1          & 73.9          & 95.10\%                                               \\
ParVTS (Ours)   & 50.7          & \textbf{81.1} & \textbf{2257.8} & 85.2          & \textbf{96.0} & \textbf{79.8} & \textbf{97.30\%}                                      \\ \hline
\rowcolor[HTML]{EFEFEF} 
                & \multicolumn{6}{c|}{\cellcolor[HTML]{EFEFEF}\textit{Dynamic Token Reduction (  $-$88.89\%)}}    &                                                       \\
FastV~\cite{fastv}           & \textbf{57.5} & 90.8          & 2108.6          & \textbf{88.5} & 75.2          & 71.4          & 91.48\%                                               \\
ParVTS (Ours)   & 45.7          & \textbf{96.5} & \textbf{2258.8} & 73.0          & \textbf{79.4} & \textbf{78.7} & \textbf{96.62\%}                            \\ \bottomrule        
\end{tabular}
}
\end{table*}

\clearpage

\section{Fine-Grained Efficiency Metrics and Analysis}
\label{sec:detail_latency}

We reported the latency, speedup ratio, GPU peak memory, and FLOPs of ParVTS for both the prefill and decoding stages, under different pruning rates and application scenarios (small batches with long responses; medium batches with medium responses; large batches with short responses) in Table \ref{tab:latency_detail}. All experiments are conducted on LLaVA-1.5-7B.

The results show that as the pruning rate increases, both the prefill-stage speedup and the decoding-stage throughput improve. The decoding-stage throughput gains are less significant because, although ParVTS reduces the KV cache size during decoding, the performance improvement is limited by memory bandwidth. The GPU peak memory and FLOPs for both the prefill and decoding stages decrease as the pruning rates increase.

Furthermore, compared to the small-batch, long-response scenario, the large-batch, short-response setting achieves more pronounced improvements in both prefill speedup and decoding throughput. This is because the prefilling phase is computation-bound, while the decoding phase is memory-bound. Under large-batch conditions, both GPU computation and memory bandwidth are heavily utilized, so the computation saved in the prefill stage and the reduced KV cache memory access in the decoding stage can be more effectively converted into acceleration. Moreover, short responses mean that the total latency is dominated by the prefill stage, where the acceleration is more pronounced, thereby further amplifying the overall inference speedup. 

This shows that ParVTS is particularly suitable for services requiring large-scale concurrency with short responses, where it can significantly improve inference efficiency and reduce memory consumption.

\begin{table*}[h]
\caption{
Stage-wise performance breakdown (Prefill vs. Decoding) of ParVTS under different pruning rates, application scenarios, and output lengths on LLaVA-1.5-7B. GM stands for GPU memory.
}
\label{tab:latency_detail}
\renewcommand{\arraystretch}{1.6}
\scriptsize
\centering
\resizebox{\textwidth}{!}{%

\begin{tabular}{c|c|c|c|c|c|c|c|c|c|c}
\toprule
\multicolumn{3}{c|}{\textbf{Settings}} &\multicolumn{2}{c|}{\textbf{Latency (s)}} & \multicolumn{4}{c|}{\textbf{GPU Memory (GB)}} & \multicolumn{2}{c}{\textbf{TFLOPs}} \\
\hline

\makecell{\textbf{Batch}\\\textbf{Size}} & \makecell{\textbf{Pruning}\\\textbf{Rates}} & \makecell{\textbf{Output}\\\textbf{Tokens}} & \makecell{\textbf{Prefill}\\\textbf{(Speedup)}} & \makecell{\textbf{Decoding}\\\textbf{(Throughput)}} & \makecell{\textbf{Prefill}\\\textbf{GM}} & \makecell{\textbf{Prefill GM}\\\textbf{- Model Size}} & \makecell{\textbf{Decoding}\\\textbf{GM}} & \makecell{\textbf{Decoding GM}\\\textbf{- Model Size}} & \makecell{\textbf{Prefill}\\\textbf{TFLOPs}} & \makecell{\textbf{Decoding}\\\textbf{TFLOPs}} 
\\ \hline

\multicolumn{11}{c}{\cellcolor[HTML]{EFEFEF} \textit{Vanilla LLaVA-1.5-7B}}   \\  

\multicolumn{1}{c|}{1}           & \multicolumn{1}{c|}{\textbackslash{}} & \multicolumn{1}{c|}{197}                   & \multicolumn{1}{c|}{0.39  (1.00$\times$)}                & \multicolumn{1}{c|}{37.05  (5.3   token/s)}      & \multicolumn{1}{c|}{14.2}            & \multicolumn{1}{c|}{0.4}                          & \multicolumn{1}{c|}{14.6}             & \multicolumn{1}{c|}{0.8}                           & \multicolumn{1}{c|}{8.9}            & \multicolumn{1}{c}{10.8}            \\ \hline

\multicolumn{1}{c|}{4}           & \multicolumn{1}{c|}{\textbackslash{}} & 114                   & \multicolumn{1}{c|}{0.73  (1.00$\times$)}                 & 22.02  (5.2    token/s)    & \multicolumn{1}{c|}{15.5}            & \multicolumn{1}{c|}{1.7}                          & \multicolumn{1}{c|}{16.7}             & 2.9                           & \multicolumn{1}{c|}{33.2}           & 15.7            \\ \hline

\multicolumn{1}{c|}{8}          & \multicolumn{1}{c|}{\textbackslash{}} & \multicolumn{1}{c|}{1}                     & \multicolumn{1}{c|}{1.24  (1.00$\times$)}                & \multicolumn{1}{c|}{0.18  (5.6   token/s)}       & \multicolumn{1}{c|}{17.3}            & \multicolumn{1}{c|}{3.5}                          & \multicolumn{1}{c|}{19.4}             & \multicolumn{1}{c|}{5.6}                           & \multicolumn{1}{c|}{67.5}           & \multicolumn{1}{c}{0.1}             \\ \hline

\multicolumn{11}{c}{\cellcolor[HTML]{EFEFEF} \textit{Small Batches with Long Responses}}   \\  

\multicolumn{1}{c|}{1}                          & \multicolumn{1}{c|}{22.22\%}          & \multicolumn{1}{c|}{196}                   & \multicolumn{1}{c|}{0.36  (1.09$\times$)}             & \multicolumn{1}{c|}{38.46  (5.1   token/s)}      & \multicolumn{1}{c|}{14.2}            & \multicolumn{1}{c|}{0.4}                          & \multicolumn{1}{c|}{14.5}             & \multicolumn{1}{c|}{0.7}                           & \multicolumn{1}{c|}{7.3}            & \multicolumn{1}{c}{9.4}             \\ \hline
\multicolumn{1}{c|}{1}                          & \multicolumn{1}{c|}{33.33\%}          & \multicolumn{1}{c|}{196}                   & \multicolumn{1}{c|}{0.36  (1.09$\times$)}             & \multicolumn{1}{c|}{38.44  (5.1   token/s)}      & \multicolumn{1}{c|}{14.1}            & \multicolumn{1}{c|}{0.3}                          & \multicolumn{1}{c|}{14.5}             & \multicolumn{1}{c|}{0.7}                           & \multicolumn{1}{c|}{6.5}            & \multicolumn{1}{c}{8.8}             \\ \hline
\multicolumn{1}{c|}{1}                          & \multicolumn{1}{c|}{50.00\%}          & \multicolumn{1}{c|}{196}                   & \multicolumn{1}{c|}{0.35  (1.15$\times$)}             & \multicolumn{1}{c|}{39.10  (5.0   token/s)}      & \multicolumn{1}{c|}{14.1}            & \multicolumn{1}{c|}{0.3}                          & \multicolumn{1}{c|}{14.4}             & \multicolumn{1}{c|}{0.6}                           & \multicolumn{1}{c|}{5.3}            & \multicolumn{1}{c}{7.8}             \\ \hline
\multicolumn{1}{c|}{1}                          & \multicolumn{1}{c|}{66.67\%}          & \multicolumn{1}{c|}{197}                   & \multicolumn{1}{c|}{0.31  (1.28$\times$)}             & \multicolumn{1}{c|}{39.13  (5.0   token/s)}      & \multicolumn{1}{c|}{14.0}            & \multicolumn{1}{c|}{0.2}                          & \multicolumn{1}{c|}{14.4}             & \multicolumn{1}{c|}{0.6}                           & \multicolumn{1}{c|}{4.2}            & \multicolumn{1}{c}{9.9}             \\ \hline
\multicolumn{1}{c|}{1}                          & \multicolumn{1}{c|}{77.78\%}          & \multicolumn{1}{c|}{195}                   & \multicolumn{1}{c|}{0.27  (1.48$\times$)}             & \multicolumn{1}{c|}{38.75  (5.0   token/s)}      & \multicolumn{1}{c|}{14.0}            & \multicolumn{1}{c|}{0.2}                          & \multicolumn{1}{c|}{14.3}             & \multicolumn{1}{c|}{0.5}                           & \multicolumn{1}{c|}{3.4}            & \multicolumn{1}{c}{8.9}             \\ \hline
\multicolumn{1}{c|}{1}                          & \multicolumn{1}{c|}{88.89\%}          & \multicolumn{1}{c|}{195}                   & \multicolumn{1}{c|}{0.27  (1.46$\times$)}             & \multicolumn{1}{c|}{36.40  (5.4   token/s)}      & \multicolumn{1}{c|}{13.9}            & \multicolumn{1}{c|}{0.1}                          & \multicolumn{1}{c|}{14.2}             & \multicolumn{1}{c|}{0.4}                           & \multicolumn{1}{c|}{2.7}            & \multicolumn{1}{c}{5.6}             \\ \hline

\multicolumn{11}{c}{\cellcolor[HTML]{EFEFEF} \textit{Medium Batches with Medium Responses}}   \\  

\multicolumn{1}{c|}{4}                          & \multicolumn{1}{c|}{22.22\%}          & 116                   & \multicolumn{1}{c|}{0.66  (1.11$\times$)}            & 22.19  (5.2    token/s)    & \multicolumn{1}{c|}{15.2}            & \multicolumn{1}{c|}{1.4}                          & \multicolumn{1}{c|}{16.2}             & 2.4                           & \multicolumn{1}{c|}{26.9}           & 14.4            \\ \hline
\multicolumn{1}{c|}{4}                          & \multicolumn{1}{c|}{33.33\%}          & 132                   & \multicolumn{1}{c|}{0.69  (1.05$\times$)}            & 30.46  (4.3    token/s)    & \multicolumn{1}{c|}{15.0}            & \multicolumn{1}{c|}{1.2}                          & \multicolumn{1}{c|}{16.0}             & 2.2                           & \multicolumn{1}{c|}{23.8}           & 16.7            \\ \hline
\multicolumn{1}{c|}{4}                          & \multicolumn{1}{c|}{50.00\%}          & 102                   & \multicolumn{1}{c|}{0.54  (1.34$\times$)}            & 20.35  (5.0    token/s)    & \multicolumn{1}{c|}{14.7}            & \multicolumn{1}{c|}{0.9}                          & \multicolumn{1}{c|}{15.6}             & 1.8                           & \multicolumn{1}{c|}{19.2}           & 9.9             \\ \hline
\multicolumn{1}{c|}{4}                          & \multicolumn{1}{c|}{66.67\%}          & 103                   & \multicolumn{1}{c|}{0.47  (1.57$\times$)}            & 19.14  (5.4    token/s)    & \multicolumn{1}{c|}{14.5}            & \multicolumn{1}{c|}{0.7}                          & \multicolumn{1}{c|}{15.2}             & 1.4                           & \multicolumn{1}{c|}{14.5}           & 9.0             \\ \hline
\multicolumn{1}{c|}{4}                          & \multicolumn{1}{c|}{77.78\%}          & 103                   & \multicolumn{1}{c|}{0.41  (1.77$\times$)}            & 18.96  (5.4    token/s)    & \multicolumn{1}{c|}{14.3}            & \multicolumn{1}{c|}{0.5}                          & \multicolumn{1}{c|}{15.0}             & 1.2                           & \multicolumn{1}{c|}{11.5}           & 8.3             \\ \hline
\multicolumn{1}{c|}{4}                          & \multicolumn{1}{c|}{88.89\%}          & 113                   & \multicolumn{1}{c|}{0.35  (2.11$\times$)}            & 21.35  (5.3    token/s)    & \multicolumn{1}{c|}{14.2}            & \multicolumn{1}{c|}{0.4}                          & \multicolumn{1}{c|}{14.8}             & 1.0                           & \multicolumn{1}{c|}{8.4}            & 8.7             \\ \hline

\multicolumn{11}{c}{\cellcolor[HTML]{EFEFEF} \textit{Large Batches with Short Responses}}   \\  

\multicolumn{1}{c|}{8}                          & \multicolumn{1}{c|}{22.22\%}          & \multicolumn{1}{c|}{1}                     & \multicolumn{1}{c|}{1.06  (1.17$\times$)}             & \multicolumn{1}{c|}{0.17  (5.7   token/s)}       & \multicolumn{1}{c|}{16.6}            & \multicolumn{1}{c|}{2.8}                          & \multicolumn{1}{c|}{18.3}             & \multicolumn{1}{c|}{4.5}                           & \multicolumn{1}{c|}{54.9}           & \multicolumn{1}{c}{0.1}             \\ \hline
\multicolumn{1}{c|}{8}                          & \multicolumn{1}{c|}{33.33\%}          & \multicolumn{1}{c|}{1}                     & \multicolumn{1}{c|}{0.99  (1.25$\times$)}             & \multicolumn{1}{c|}{0.18  (5.5   token/s)}       & \multicolumn{1}{c|}{16.3}            & \multicolumn{1}{c|}{2.5}                          & \multicolumn{1}{c|}{17.8}             & \multicolumn{1}{c|}{4.0}                           & \multicolumn{1}{c|}{48.7}           & \multicolumn{1}{c}{0.1}             \\ \hline
\multicolumn{1}{c|}{8}                          & \multicolumn{1}{c|}{50.00\%}          & \multicolumn{1}{c|}{1}                     & \multicolumn{1}{c|}{0.81  (1.53$\times$)}             & \multicolumn{1}{c|}{0.17  (5.8   token/s)}       & \multicolumn{1}{c|}{15.7}            & \multicolumn{1}{c|}{2.0}                          & \multicolumn{1}{c|}{17.0}             & \multicolumn{1}{c|}{3.2}                           & \multicolumn{1}{c|}{39.4}           & \multicolumn{1}{c}{0.1}             \\ \hline
\multicolumn{1}{c|}{8}                          & \multicolumn{1}{c|}{66.67\%}          & \multicolumn{1}{c|}{1}                     & \multicolumn{1}{c|}{0.71  (1.74$\times$)}             & \multicolumn{1}{c|}{0.17  (5.9   token/s)}       & \multicolumn{1}{c|}{15.2}            & \multicolumn{1}{c|}{1.5}                          & \multicolumn{1}{c|}{16.1}             & \multicolumn{1}{c|}{2.3}                           & \multicolumn{1}{c|}{30.1}           & \multicolumn{1}{c}{0.1}             \\ \hline
\multicolumn{1}{c|}{8}                          & \multicolumn{1}{c|}{77.78\%}          & \multicolumn{1}{c|}{1}                     & \multicolumn{1}{c|}{0.58  (2.15$\times$)}             & \multicolumn{1}{c|}{0.18  (5.7   token/s)}       & \multicolumn{1}{c|}{14.9}            & \multicolumn{1}{c|}{1.1}                          & \multicolumn{1}{c|}{15.6}             & \multicolumn{1}{c|}{1.8}                           & \multicolumn{1}{c|}{24.0}           & \multicolumn{1}{c}{0.1}             \\ \hline
\multicolumn{1}{c|}{8}                          & \multicolumn{1}{c|}{88.89\%}          & \multicolumn{1}{c|}{1}                     & \multicolumn{1}{c|}{0.52  (2.40$\times$)}              & \multicolumn{1}{c|}{0.16  (6.1   token/s)}       & \multicolumn{1}{c|}{14.6}            & \multicolumn{1}{c|}{0.8}                          & \multicolumn{1}{c|}{15.1}             & \multicolumn{1}{c|}{1.3}                           & \multicolumn{1}{c|}{17.9}           & \multicolumn{1}{c}{0.1}

\\ \bottomrule
\end{tabular}
}
\end{table*}

\section{Theoretical Speedup Model}
\label{sec:theoretical}

We establish a theoretical model for the speedup ratio, with the key variables defined as follows:

\begin{itemize}
\item $p$: pruning rate of visual tokens.
\item $n$: migration depth.
\item $N$: total number of transformer layers in the LLM.
\item $L$: input token length.
\item $L_{\text{text}}$: length of system and question tokens.
\item $L_{\text{img}}$: total length of visual tokens.
\item $L_{\text{sub}}$: length of subject tokens, where $L_{\text{sub}} = (1 - p) \times L_{\text{img}}$.
\item $M$: output token length during autoregressive decoding.
\item $d$: hidden state dimension.
\item $m$: intermediate dimension of the feed-forward network (FFN).
\end{itemize}

Following FastV~\cite{fastv},  we consider the computation of the multi-head attention (MHA) and feed-forward network (FFN) module in the FLOPs estimation. 
For a single transformer layer, the total FLOPs can be estimated as:
\begin{align*}
    FLOPs_{\text{layer}}=4d^2L+2dL^2+2mdL.
\end{align*}

\noindent \textbf{Vanilla MLLM.}
For a vanilla MLLM, the FLOPs in the prefill stage are given by: 

\begin{align*}
FLOPs_{\text{(v, prefill)}}
    =N\cdot(4d^2L+2dL^2+2mdL).
\end{align*}
During the decoding stage, denote the KV cache length when generating the $i$-th token as $S_i$, where $S_i = L + i - 1$. The FLOPs in the decoding stage can be written as:
\begin{align*}
FLOPs_{\text{(v, decoding)}}=N \cdot  \sum_{i=1}^{M} ( 4d^2 + 2dS_i   + 2md  ).
\end{align*}
Substituting $S_i = L + i - 1$, we obtain: 
\begin{align*}
FLOPs_{\text{(v, decoding)}}
    =N \cdot  \sum_{i=1}^{M} \big( 4d^2 + 2d(L+i-1)   + 2md  \big),
\end{align*}
which simplifies to: 
\begin{align*}
FLOPs_{\text{(v, decoding)}}= N \cdot M \big(  4d^2 + 2d(L+\frac{M-1}{2} ) +2md \big).
\end{align*}

\noindent \textbf{ParVTS.}
For ParVTS, the FLOPs in the prefill stage consist of two parts:

1) For the first $n$ layers, the FLOPs can be expressed as: 
\begin{align*}
FLOPs_1 = n\cdot(4Ld^2+2L^2d+2Lmd).
\end{align*}

2) For layers $n+1$ to $N$, the token length changes from $L$ to $L_{\text{text}} + L_{\text{sub}}$, and the FLOPs are computed as: 
\begin{align*}
& FLOPs_2 = (N-n) \Big( 4d^2 \big(L_{\text{text}}+L_{\text{sub}}  \big)  + 2d \big(L_{\text{text}}+L_{\text{sub}}  \big)^2 
 +2md\big(L_{\text{text}}+L_{\text{sub}}  \big) \Big).
\end{align*}

By summing $FLOPs_1$ and $FLOPs_2$, and substituting $L_{\text{sub}} = (1-p)L_{\text{img}}$, we obtain the FLOPs of ParVTS in the prefill stage as: 
\begin{align*}
    &FLOPs_{\text{(p, prefill)}} = n\cdot(4Ld^2+2L^2d+2Lmd) 
    +(N-n) \Big( 4d^2 \big(L_{\text{text}}+(1-p)L_{\text{img}}  \big) \\
    & \quad  + 2d \big(L_{\text{text}}+(1-p)L_{\text{img}}  \big)^2 +2md\big(L_{\text{text}}+(1-p)L_{\text{img}}  \big) \Big).
    \end{align*}

Similarly, the FLOPs of ParVTS in the decoding stage are given by: 
\begin{align*}
    & FLOPs_{\text{(p, decoding)}}=N\cdot M\cdot \Big (  4d^2+2md 
     +2d \big( \frac{M-1}{2}
    +L_{\text{text}}+(1-p)L_{\text{img}} \big) \Big ).
\end{align*}

\noindent \textbf{Speedup Ratio.}
Assuming the hardware computation rate is $R_{GPU}$ (FLOPs/s), we then define the speedup ratio as:  
\begin{align*}
\rho = \frac{T_{\text{vanilla}}}{T_{\text{ParVTS}}} = \frac{T_{\text{vanilla}} \cdot R_{\text{GPU}}}{T_{\text{ParVTS}} \cdot R_{\text{GPU}}} = \frac{FLOPs_{\text{(v)}} }{FLOPs_{\text{(p)}} }. 
\end{align*}
 Then the speedup ratio in the prefill stage is therefore: 
\begin{align*}
\rho_{\text{prefill}} = \frac{FLOPs_{\text{(v, prefill)}}}{FLOPs_{\text{(p, prefill)}}} 
 \end{align*}
 Since the formula is too long, we will not perform further substitution here.

The speedup ratio in the decoding stage is: 
\begin{align*}
    &\rho_{\text{decoding}} = \frac{FLOPs_{\text{(v, decoding)}}}{FLOPs_{\text{(p, decoding)}}} 
    = \frac{  2d +m + (L+\frac{M-1}{2} )   }{  2d+m+ \big( \frac{M-1}{2} +L_{\text{text}}+(1-p)L_{\text{img}} \big) }.
\end{align*}

\noindent \textbf{Analysis of Speedup Factors.}
We analyze the speedup ratios in both prefill and decoding stages with respect to three key variables: 
\begin{itemize}
    \item Pruning rate $p$;
    \item Migration depth $n$;
    \item Output token length $M$.
\end{itemize}

\noindent \textbf{\textit{Pruning rate $p$.}}
In both the prefill and decoding phases, the pruning rate $p$ is positively correlated with the speedup ratio $\rho$. However, in practice, the acceleration during decoding is often limited by memory bandwidth, resulting in a smaller effective speedup compared to the prefill stage under the same pruning rate.

\noindent \textbf{\textit{Migration depth $n$.}}  
   For the prefill stage, increasing $n$ decreases the overall speedup since more layers operate before pruning is applied, yet $\rho_{\text{prefill}}$ always remains above~1. From the expression of $\rho_{\text{decoding}}$, it is evident that $n$ does not affect the decoding speedup, implying that migration depth primarily influences the prefill latency.

\noindent \textbf{\textit{Output token length $M$.}}  
    The output length $M$ has no influence on the prefill-stage speedup. 
    Differentiating $\rho_{\text{decoding}}$ with respect to $M$ yields:
    $$\frac{d\rho_{\text{decoding}}}{dM} = \frac{\mathcal{N}}{\mathcal{D}},$$
    where the numerator is:
    $$\mathcal{N} = 6d^2 \big(L_{\text{text}} + (1-p)L_{\text{img}} - L\big),$$
    and the denominator is: 
    $$\mathcal{D} = \big(\,4d^2 + 2md + 2d\big(\tfrac{M-1}{2} + L_{\text{text}} + (1-p)L_{\text{img}}\big)\,\big)^2.$$
    Evidently, the numerator $\mathcal{N}$ simplifies to a negative value since $L_{\text{text}} + (1-p)L_{\text{img}} - L = -p L_{\text{img}} < 0$. The denominator $\mathcal{D}$ is always positive, as it is a squared term. Thus, $\frac{d\rho_{\text{decoding}}}{dM} < 0$, which means that as the output token length $M$ increases, the speedup ratio in the decoding phase decreases.

\section{Qualitative Results on Downstream Tasks}
\label{sec:lisa}
\begin{figure*}[!]
    \centering
    \includegraphics[width=\textwidth]{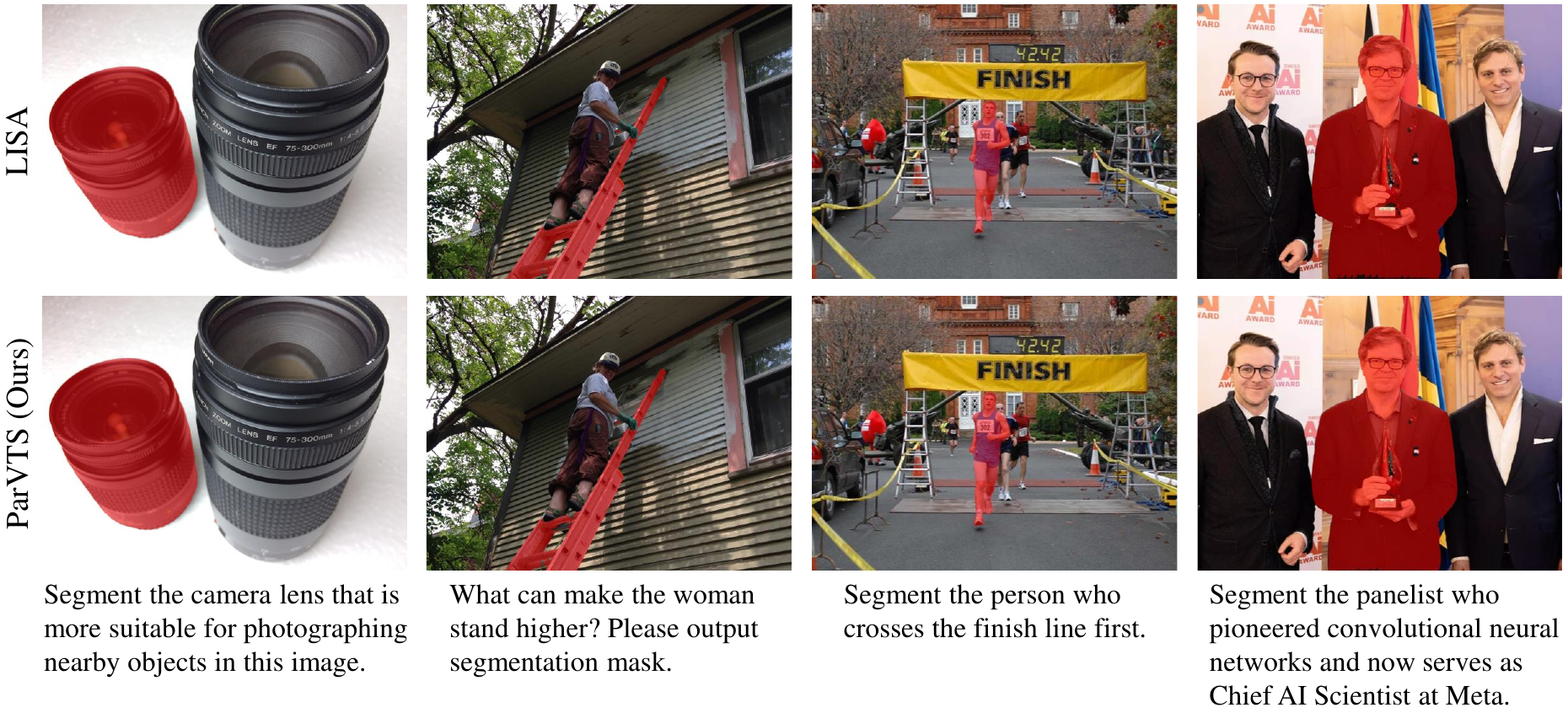}
    \caption{
    ParVTS performance on the LISA~\cite{lai2024lisa} segmentation task.
    }
    \label{fig:lisa}
\end{figure*}

To demonstrate ParVTS's ability to preserve fine-grained visual reasoning and robust segmentation fidelity, we present qualitative examples on the LISA segmentation task in Figure\,\ref{fig:lisa}. 
The results illustrate that ParVTS successfully maintains the original model's precise segmentation capability under significant token reduction. Specifically, ParVTS accurately handles complex, instruction-based queries that demand detailed visual grounding and differentiating skills. 

For instance, it correctly distinguishes between two visually similar camera lenses based on a functional description ("more suitable for photographing nearby objects"). Furthermore, in challenging scenarios, ParVTS retains the ability to segment based on temporal order (identifying "the person who crosses the finish line first") and external knowledge (locating "a specific scientist" based on their professional title). 
Crucially, the segmentation masks produced by ParVTS are highly consistent with the original LISA outputs, validating its effectiveness in selectively pruning tokens while preserving the information necessary for intricate visual comprehension and precise localization.

\section{Additional Analysis of Sequential Scheduling Variants}
\label{appendix:scheduling_variants}

Table\,\ref{tab:subfs_detail} provides a detailed comparison of Subject-First Scheduling (Sub-FS) and Non-Subject-First Scheduling (Non-Sub-FS) under various migration depths $n$. 

For Sub-FS, a small $n$ limits the exposure of subject tokens to early transformer layers, resulting in incomplete semantic transfer into the question token representations and degraded performance. Moreover, when $n$ is small, the large number of non-subject tokens, and their prolonged activity across many layers, leads to slow inference, especially under high pruning rates. 
As $n$ increases, performance improves due to more sufficient early-stage information migration. Latency—especially under high pruning rates—also decreases, as non-subject tokens, which are more numerous in this setting, are introduced later and remain active for fewer layers. However, deferring their introduction limits their contribution to information transfer, resulting in diminishing performance gains.

For Non-Sub-FS, a large $n$ also leads to slower inference, as non-subject tokens are retained for more layers before being discarded. When $n$ is small, performance remains acceptable, but shallow representation and semantic mismatch (as discussed in Sec.\,\ref{sec:when_and_how}) still limit its effectiveness. 

Overall, ParVTS achieves superior performance-efficiency trade-offs by enabling both token groups to participate early and eliminating the need for abrupt switching.

\section{Additional Ablation Results of Fusion Weights $\alpha$ and $\beta$}
\label{appendix:alpha_beta_ablation}
We provide additional ablation results of the fusion weights $\alpha$ and $\beta$ under different pruning rates. As shown in Table\,\ref{tab:ablation_alpha_full}, setting $\alpha = 0.5$ and $\beta = 0.5$ consistently achieves the best performance across all pruning levels.

\section{Visualization of [CLS] Token Attention and Qualitative Results}
\label{sec:visual_cls_token}
\begin{figure*}[!]
    \centering
    \includegraphics[width=0.9\textwidth]{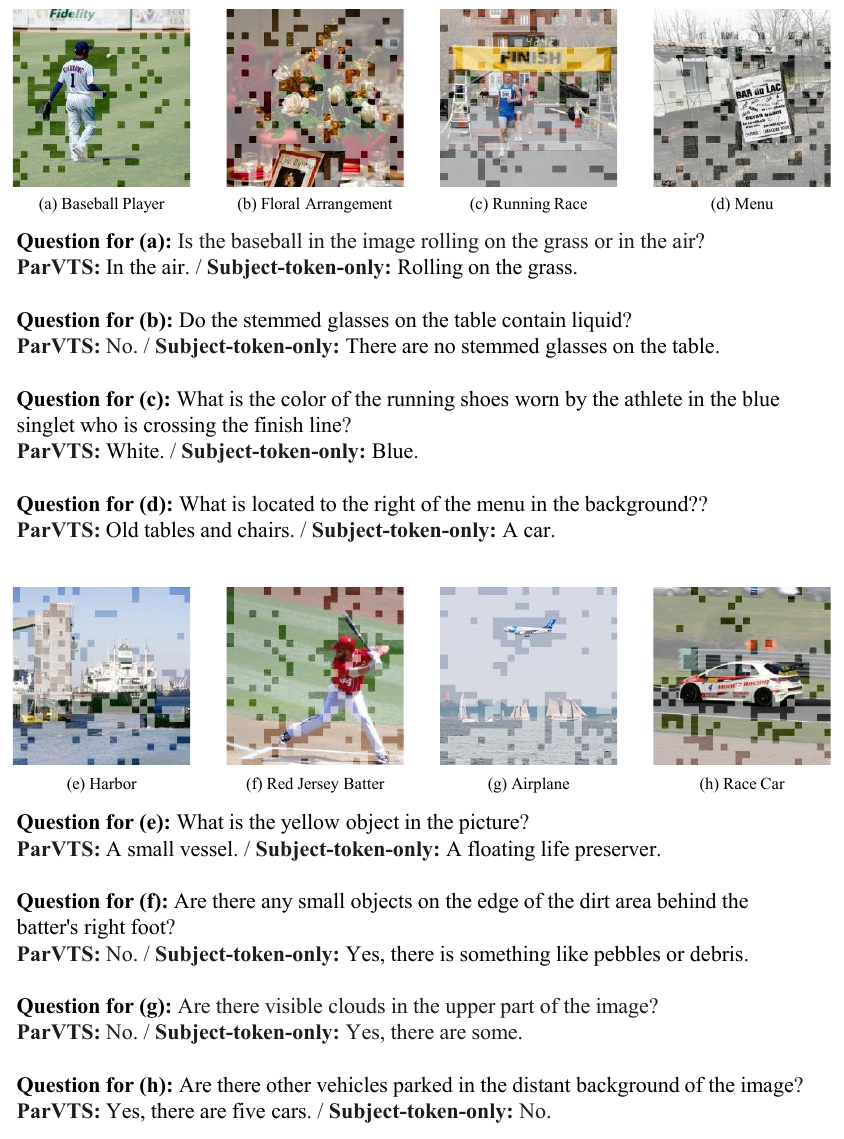}
    \caption{
    Visualization of [CLS] token attention and qualitative results of ParVTS on LLaVA-1.5-7B.
    }
    \label{fig:inference_example}
\end{figure*}

We present qualitative examples of the model’s inference on real-world scenarios to illustrate ParVTS's visual reasoning capabilities. As shown in Figure\,\ref{fig:inference_example}, ParVTS correctly answers questions targeting non-subject areas, whereas subject-token-only methods (e.g., FastV~\cite{fastv}) tend to produce incorrect responses.

For instance, in scene (c), the query asks for the color of the running shoes worn by the athlete, a detailed attribute that the subject-token-only method misses, incorrectly responding "Blue" while ParVTS correctly identifies the color as "White." 
These failures underscore the necessity of our approach, as ParVTS, by scheduling high-value non-subject tokens, maintains the comprehensive visual context required for intricate global visual grounding.

In addition, we visualize the regions identified by [CLS] token attention, where subject tokens are displayed as clean areas and non-subject tokens are overlaid with a 50\% opacity mask. The visualizations show that the [CLS]-based attention effectively localizes foreground objects, confirming the reliability of using [CLS]-token scores to distinguish subject from background regions.

\begin{table}[h]
\renewcommand{\arraystretch}{1.6}
\caption{Performance and latency comparison of token scheduling strategies under different token budgets.}
\label{tab:subfs_detail}
\centering
\resizebox{\textwidth}{!}{%
\begin{tabular}{c|ccccccc|c|c|c}
\toprule
\multicolumn{1}{c|}{}                                  & \multicolumn{8}{c|}{\textbf{Metrics}}                                                                                                                                                                                                                                                                              & \multicolumn{2}{c}{\textbf{Latency}}                                                                   \\ \cline{2-11} 
\multicolumn{1}{c|}{\multirow{-2}{*}{\textbf{Method}}} & \multicolumn{1}{c}{\textbf{GQA}} & \multicolumn{1}{c}{\textbf{MMB}} & \multicolumn{1}{c}{\textbf{MME}} & \multicolumn{1}{c}{\textbf{POPE}} & \multicolumn{1}{c}{\textbf{SQA}} & \multicolumn{1}{c}{\textbf{VizWiz}} & \multicolumn{1}{c|}{\textbf{OCRBench}} & \multicolumn{1}{c|}{\textbf{Avg. $\uparrow$}} & \multicolumn{1}{l|}{\textbf{\makecell{Time / Sample \\  (ms) $\downarrow$}}} & \multicolumn{1}{c}{\textbf{Speedup$\uparrow$}} \\ \hline
\rowcolor[HTML]{EFEFEF} 
LLaVA-1.5-7B                                            & \multicolumn{7}{c|}{\cellcolor[HTML]{EFEFEF}\textit{All tokens (Dynamic,   100.00\%)}}                                                                                                                                                                              &                                               &                                                                 &                                       \\
Vanilla                                                 & 61.9                              & 64.1                              & 1866.1                            & 85.9                               & 69.6                              & 54.3                                 & 313.0                                  & 100.00\%                                      & 285.17                                                          & 1.00$\times$                          \\ \hline
\rowcolor[HTML]{EFEFEF} 
  & \multicolumn{7}{c|}{\cellcolor[HTML]{EFEFEF}\textit{Dynamic Token Reduction (  $-$66.67\%)}}                                                                                                                                                                       &                                               &                                                                 &                                       \\
Sub-FS ($n=3$)                                          & 57.1                              & 52.7                              & 1531.5                            & 76.7                               & 65.8                              & 53.2                                 & 119.0                                  & 77.77\%                                       & 235.47                                                          & 1.21$\times$                          \\
Sub-FS   ($n=16$)                                       & 55.3                              & 63.7                              & 1850.1                            & 84.4                               & 68.5                              & 52.2                                 & 79.0                                   & 89.58\%                                       & 216.51                                                          & 1.32$\times$                          \\
Sub-FS   ($n=20$)                                       & 58.5                              & 63.8                     & 1843.1                            & 84.6                               & 68.3                              & 54.3                                 & 254.0                                  & 96.48\%                                       & 212.30                                                          & 1.34$\times$                          \\
Non-Sub-FS   ($n=3$)                                    & 59.5                              & 63.2                              & 1825.8                            & 83.6                               & 68.0                              & 53.6                                 & 293.0                                  & 97.28\%                                       & 190.40                                                          & 1.50$\times$                          \\
Non-Sub-FS   ($n=16$)                                   & 55.8                              & 52.7                              & 1518.9                            & 76.9                               & 66.1                              & 53.7                                 & 170.0                                  & 79.28\%                                       & 215.25                                                          & 1.32$\times$                          \\
Non-Sub-FS   ($n=20$)                                   & 56.5                              & 52.6                              & 1508.2                            & 77.1                               & 66.2                              & 53.4                                 & 139.0                                  & 77.65\%                                       & 221.15                                                          & 1.29$\times$                          \\
ParVTS   (Ours)  ($n=3$)                                & 59.7                     & 63.6                              & 1831.9                            & 84.6                               & 68.5                              & 55.2                                 & 300.0                                  & 97.95\%                                       & 205.14                                                          & 1.39$\times$                          \\ \hline
\rowcolor[HTML]{EFEFEF} 
 & \multicolumn{7}{c|}{\cellcolor[HTML]{EFEFEF}\textit{Dynamic Token Reduction (  $-$77.78\%)}}                                                                                                                                                                       &                                               &                                                                 &                                       \\
Sub-FS ($n=3$)                                          & 58.8                              & 57.6                              & 1651.5                            & 80.3                               & 66.9                              & 53.7                                 & 168.0                                  & 84.96\%                                       & 239.68                                                          & 1.19$\times$                          \\
Sub-FS   ($n=16$)                                       & 54.4                              & 63.2                              & 1802.9                            & 81.6                               & 68.6                              & 52.7                                 & 106.0                                  & 88.64\%                                       & 208.51                                                          & 1.37$\times$                          \\
Sub-FS   ($n=20$)                                       & 57.3                              & 63.4                              & 1799.3                            & 82.0                               & 68.3                              & 54.1                                 & 244.0                                  & 94.17\%                                       & 199.24                                                          & 1.43$\times$                          \\
Non-Sub-FS   ($n=3$)                                    & 58.2                              & 62.9                              & 1766.2                            & 81.0                               & 68.3                              & 54.1                                 & 291.0                                  & 94.69\%                                       & 173.55                                                          & 1.64$\times$                          \\
Non-Sub-FS   ($n=16$)                                   & 57.0                              & 57.0                              & 1663.9                            & 80.6                               & 66.4                              & 53.9                                 & 193.0                                  & 86.35\%                                       & 205.14                                                          & 1.39$\times$                          \\
Non-Sub-FS   ($n=20$)                                   & 57.9                              & 57.0                              & 1665.0                            & 81.1                               & 66.2                              & 53.7                                 & 176.0                                  & 85.76\%                                       & 214.41                                                          & 1.33$\times$                          \\
ParVTS   (Ours)  ($n=3$)                                & 58.2                              & 63.1                              & 1789.2                            & 82.0                               & 68.6                              & 55.3                                 & 297.0                                  & 95.96\%                                       & 173.13                                                          & 1.65$\times$                          \\ \hline
\rowcolor[HTML]{EFEFEF} 
& \multicolumn{7}{c|}{\cellcolor[HTML]{EFEFEF}\textit{Dynamic Token Reduction (  $-$88.89\%)}}                                                                                                                                                                       &                                               &                                                                 &                                       \\
Sub-FS ($n=3$)                                          & 54.6                              & 61.7                              & 1636.8                            & 79.3                               & 66.5                              & 53.2                                 & 210.0                                  & 85.97\%                                       & 240.10                                                          & 1.19$\times$                          \\
Sub-FS   ($n=16$)                                       & 53.0                              & 61.6                              & 1742.7                            & 76.4                               & 68.6                              & 53.4                                 & 124.0                                  & 86.67\%                                       & 200.93                                                          & 1.42$\times$                          \\
Sub-FS   ($n=20$)                                       & 54.6                              & 60.7                              & 1717.0                            & 76.7                               & 68.5                              & 54.4                                 & 226.0                                  & 89.77\%                                       & 192.08                                                          & 1.48$\times$                          \\
Non-Sub-FS   ($n=3$)                                    & 55.2                              & 61.2                              & 1671.9                            & 76.0                               & 68.7                              & 54.4                                 & 271.0                                  & 89.80\%                                       & 165.54                                                          & 1.72$\times$                          \\
Non-Sub-FS   ($n=16$)                                   & 57.5                              & 61.6                              & 1719.2                            & 83.9                               & 66.7                              & 53.7                                 & 194.0                                  & 88.93\%                                       & 192.50                                                          & 1.48$\times$                          \\
Non-Sub-FS   ($n=20$)                                   & 59.4                              & 61.5                              & 1711.0                            & 82.6                               & 66.9                              & 54.0                                 & 195.0                                  & 88.68\%                                       & 211.46                                                          & 1.35$\times$                          \\
ParVTS   (Ours)  ($n=3$)                                & 55.2                              & 61.3                              & 1739.4                            & 76.9                               & 68.9                              & 55.1                                 & 282.0                                  & 92.91\%                                       & 172.70                                                          & 1.65$\times$   \\ \bottomrule                      
\end{tabular}
}
\end{table}

\begin{table*}[]
\renewcommand{\arraystretch}{1.0}
\caption{Ablation studies of fusion weights $
\alpha$ and $\beta$ on LLaVA-1.5-7B under different vision token budgets.}
\label{tab:ablation_alpha_full}
\centering
\resizebox{\textwidth}{!}{%
\begin{tabular}{c|ccccccc|c}
\toprule
\multicolumn{1}{c|}{\textbf{Method}}                    & \multicolumn{1}{c}{\textbf{GQA}}    & \multicolumn{1}{c}{\textbf{MMB}}    & \multicolumn{1}{c}{\textbf{MME}}      & \multicolumn{1}{c}{\textbf{POPE}}   & \multicolumn{1}{c}{\textbf{SQA}}    & \multicolumn{1}{c}{\textbf{VizWiz}} & \multicolumn{1}{c|}{\textbf{OCRBench}} & \multicolumn{1}{c}{\textbf{Avg. $\uparrow$}} \\ \hline
\rowcolor[HTML]{EFEFEF} 
LLaVA-1.5-7B~\cite{liu2023llava}                                             & \multicolumn{7}{c|}{\cellcolor[HTML]{EFEFEF}\textit{All tokens (576,   100.00\%)}}                                                                                                                                                                                                  &                                               \\
Vanilla                                                  & 61.9                                 & 64.1                                 & 1866.1                                 & 85.9                                 & 69.6                                 & 54.3                                 & 313.0                                  & 100.00\%                                      \\ \hline
\rowcolor[HTML]{EFEFEF} 

& \multicolumn{7}{c}{\cellcolor[HTML]{EFEFEF}\textit{Token Reduction (448,   $-$22.22\%)}}                                                                                                                                                                                           &                                               \\
$\alpha$=0.0   $\beta$=1.0                               & 61.1                                 & 63.3                                 & 1846.9                                 & 86.0                                 & 68.4                                 & 54.1                                 & 311.0                                  & 99.04\%                                       \\
$\alpha$=0.1   $\beta$=0.9                               & 61.5                                 & 63.5                                 & 1846.6                                 & 86.4                                 & 68.7                                 & 54.3                                 & 310.0                                  & 99.04\%                                       \\
$\alpha$=0.3   $\beta$=0.7                               & 61.4                                 & 63.7                                 & 1848.8                                 & 86.3                                 & 68.5                                 & 54.0                                 & 310.0                                  & 99.11\%                                       \\
{ $\alpha$=0.5   $\beta$=0.5 (Ours)} & {\color[HTML]{333333} \textbf{61.8}} & {\color[HTML]{333333} \textbf{64.0}} & {\color[HTML]{333333} \textbf{1853.2}} & {\color[HTML]{333333} \textbf{86.8}} & {\color[HTML]{333333} \textbf{69.0}} & {\color[HTML]{333333} \textbf{54.7}} & {\color[HTML]{333333} \textbf{315.0}}  & {\color[HTML]{333333} \textbf{99.58\%}}       \\
$\alpha$=0.7   $\beta$=0.3                               & 60.8                                 & 63.6                                 & 1848.9                                 & 86.4                                 & 68.1                                 & 54.0                                 & 310.0                                  & 99.07\%                                       \\
$\alpha$=0.9  $\beta$=0.1                                & 60.0                                 & 61.7                                 & 1841.7                                 & 83.7                                 & 66.1                                 & 52.0                                 & 308.0                                  & 98.33\%                                       \\
$\alpha$=1.0  $\beta$=0.0                                & 59.5                                 & 61.5                                 & 1837.4                                 & 83.4                                 & 66.4                                 & 49.9                                 & 308.0                                  & 98.05\%                                       \\ \hline
\rowcolor[HTML]{EFEFEF} 

 & \multicolumn{7}{c}{\cellcolor[HTML]{EFEFEF}\textit{Token Reduction (384,   $-$33.33\%)}}                                                                                                                                                                                           &                                               \\
$\alpha$=0.0   $\beta$=1.0                               & 61.1                                 & 62.8                                 & 1828.4                                 & 86.5                                 & 68.7                                 & 53.6                                 & 310.0                                  & 98.25\%                                       \\
$\alpha$=0.1   $\beta$=0.9                               & 61.6                                 & 63.6                                 & 1829.0                                 & 86.5                                 & 68.9                                 & 54.0                                 & 310.0                                  & 98.35\%                                       \\
$\alpha$=0.3   $\beta$=0.7                               & 61.3                                 & 63.5                                 & 1828.7                                 & 85.7                                 & 68.7                                 & 54.0                                 & 310.0                                  & 98.28\%                                       \\
{ $\alpha$=0.5   $\beta$=0.5 (Ours)} & {\color[HTML]{333333} \textbf{61.8}} & {\color[HTML]{333333} \textbf{64.1}} & {\color[HTML]{333333} \textbf{1835.4}} & {\color[HTML]{333333} \textbf{86.9}} & {\color[HTML]{333333} \textbf{69.2}} & {\color[HTML]{333333} \textbf{54.5}} & {\color[HTML]{333333} \textbf{314.0}}  & {\color[HTML]{333333} \textbf{98.84\%}}       \\
$\alpha$=0.7   $\beta$=0.3                               & 61.0                                 & 63.7                                 & 1832.7                                 & 86.4                                 & 68.9                                 & 54.1                                 & 313.0                                  & 98.59\%                                       \\
$\alpha$=0.9  $\beta$=0.1                                & 60.0                                 & 63.5                                 & 1829.0                                 & 86.3                                 & 68.6                                 & 54.1                                 & 313.0                                  & 98.38\%                                       \\
$\alpha$=1.0  $\beta$=0.0                                & 59.9                                 & 63.3                                 & 1828.8                                 & 86.1                                 & 68.5                                 & 53.5                                 & 312.0                                  & 98.29\%                                       \\ \hline
\rowcolor[HTML]{EFEFEF}  

& \multicolumn{7}{c}{\cellcolor[HTML]{EFEFEF}\textit{Token Reduction (Dynamic,   $-$66.67\%)}}                                                                                                                                                                                       &                                               \\
$\alpha$=0.0   $\beta$=1.0                               & 59.3                                 & 63.3                                 & 1829.7                                 & 84.4                                 & 68.2                                 & 55.0                                 & \textbf{303.0}                         & 97.93\%                                       \\
$\alpha$=0.1   $\beta$=0.9                               & 59.2                                 & 63.3                                 & 1830.7                                 & 84.4                                 & 67.4                                 & \textbf{55.3}                        & 302.0                                  & 97.90\%                                       \\
$\alpha$=0.3   $\beta$=0.7                               & \textbf{59.8}                        & \textbf{63.7}                        & 1831.4                                 & \textbf{84.6}                        & 67.5                                 & 55.1                                 & \textbf{303.0}                         & 98.01\%                                       \\
{ $\alpha$=0.5   $\beta$=0.5 (Ours)} & {\color[HTML]{333333} 59.7}          & {\color[HTML]{333333} 63.6}          & {\color[HTML]{333333} \textbf{1831.9}} & {\color[HTML]{333333} \textbf{84.6}} & {\color[HTML]{333333} \textbf{68.5}} & {\color[HTML]{333333} \textbf{55.3}} & {\color[HTML]{333333} 302.0}           & {\color[HTML]{333333} \textbf{98.03\%}}       \\
$\alpha$=0.7   $\beta$=0.3                               & 59.4                                 & 63.0                                 & 1825.9                                 & 84.3                                 & 68.3                                 & 55.1                                 & 300.0                                  & 97.65\%                                       \\
$\alpha$=0.9  $\beta$=0.1                                & 59.2                                 & 63.0                                 & 1825.6                                 & 83.5                                 & 68.0                                 & 55.1                                 & 300.0                                  & 97.59\%                                       \\
$\alpha$=1.0  $\beta$=0.0                                & 59.3                                 & 63.0                                 & 1825.1                                 & 84.0                                 & 67.8                                 & 55.0                                 & 298.0                                  & 97.50\%                                       \\ \hline
\rowcolor[HTML]{EFEFEF} 

& \multicolumn{7}{c}{\cellcolor[HTML]{EFEFEF}\textit{Token Reduction (128,   $-$77.78\%)}}                                                                                                                                                                                           &                                               \\
$\alpha$=0.0   $\beta$=1.0                               & 57.8                                 & 62.6                                 & 1782.8                                 & 81.2                                 & 68.0                                 & 55.1                                 & 292.0                                  & 95.40\%                                       \\
$\alpha$=0.1   $\beta$=0.9                               & 58.0                                 & 62.7                                 & 1783.5                                 & 81.3                                 & 68.0                                 & 55.1                                 & 292.0                                  & 95.45\%                                       \\
$\alpha$=0.3   $\beta$=0.7                               & 57.9                                 & 62.7                                 & 1784.0                                 & 81.4                                 & 67.9                                 & 55.1                                 & 292.0                                  & 95.46\%                                       \\
{ $\alpha$=0.5   $\beta$=0.5 (Ours)} & { \textbf{58.2}} & { \textbf{63.1}} & { \textbf{1789.2}} & { \textbf{82.0}} & { \textbf{68.6}} & { \textbf{55.3}} & { \textbf{296.0}}  & { \textbf{95.92\%}}       \\
$\alpha$=0.7   $\beta$=0.3                               & 57.9                                 & 62.7                                 & 1784.1                                 & 81.5                                 & 68.1                                 & 55.0                                 & 292.0                                  & 95.48\%                                       \\
$\alpha$=0.9  $\beta$=0.1                                & 57.9                                 & 62.7                                 & 1784.6                                 & 81.7                                 & 67.9                                 & 55.0                                 & 293.0                                  & 95.53\%                                       \\
$\alpha$=1.0  $\beta$=0.0                                & 57.9                                 & 62.6                                 & 1784.5                                 & 81.7                                 & 68.0                                 & 55.0                                 & 293.0                                  & 95.53\%                                       \\ \hline
\rowcolor[HTML]{EFEFEF} 

& \multicolumn{7}{c}{\cellcolor[HTML]{EFEFEF}\textit{Token Reduction (64,   $-$88.89\%)}}                                                                                                                                                                                            &                                               \\
$\alpha$=0.0   $\beta$=1.0                               & 53.1                                 & 58.8                                 & 1723.5                                 & 75.2                                 & 66.3                                 & 52.6                                 & 273.0                                  & 91.55\%                                       \\
$\alpha$=0.1   $\beta$=0.9                               & 52.9                                 & 58.9                                 & 1724.3                                 & 75.3                                 & 66.3                                 & 52.7                                 & 273.0                                  & 91.58\%                                       \\
$\alpha$=0.3   $\beta$=0.7                               & 54.9                                 & 60.6                                 & 1732.9                                 & 76.4                                 & 68.1                                 & 54.9                                 & 276.0                                  & 92.39\%                                       \\
{ $\alpha$=0.5   $\beta$=0.5 (Ours)} & {\color[HTML]{333333} \textbf{55.2}} & {\color[HTML]{333333} \textbf{61.3}} & {\color[HTML]{333333} \textbf{1739.4}} & {\color[HTML]{333333} \textbf{76.9}} & {\color[HTML]{333333} \textbf{68.9}} & {\color[HTML]{333333} \textbf{55.1}} & {\color[HTML]{333333} \textbf{280.0}}  & {\color[HTML]{333333} \textbf{92.91\%}}       \\
$\alpha$=0.7   $\beta$=0.3                               & 54.8                                 & 60.8                                 & 1733.5                                 & 76.3                                 & 68.7                                 & 54.6                                 & 276.0                                  & 92.43\%                                       \\
$\alpha$=0.9  $\beta$=0.1                                & 55.1                                 & 61.8                                 & 1738.1                                 & 76.6                                 & 68.7                                 & 54.6                                 & 279.0                                  & 92.79\%                                       \\
$\alpha$=1.0  $\beta$=0.0                                & 55.1                                 & 60.8                                 & 1734.7                                 & 76.5                                 & 68.4                                 & 54.6                                 & 277.0                                  & 92.52\%                                      
  \\ \bottomrule        
\end{tabular}
}
\end{table*}

\end{document}